%% file: camera_ready.tex
\def\year{2021}\relax
\documentclass[letterpaper]{article} % DO NOT CHANGE THIS
\usepackage{aaai21}  % DO NOT CHANGE THIS
\usepackage{times}  % DO NOT CHANGE THIS
\usepackage{helvet} % DO NOT CHANGE THIS
\usepackage{courier}  % DO NOT CHANGE THIS
\usepackage[hyphens]{url}  % DO NOT CHANGE THIS
\usepackage{graphicx} % DO NOT CHANGE THIS
\usepackage{natbib}  % DO NOT CHANGE THIS AND DO NOT ADD ANY OPTIONS TO IT
\usepackage{caption} % DO NOT CHANGE THIS AND DO NOT ADD ANY OPTIONS TO IT
\usepackage{color}
\usepackage{graphicx}
\usepackage{lineno}
\usepackage{amsmath,amssymb} % define this before the line numbering.
\usepackage{subfiles}
\usepackage{amsfonts}

\title{Graph and Temporal Convolutional Networks for 3D Multi-person\\ Pose Estimation in Monocular Videos }
\author {Yu Cheng,\textsuperscript{\rm 1} Bo Wang,\textsuperscript{\rm 2} Bo Yang,\textsuperscript{\rm 2} Robby T. Tan\textsuperscript{\rm 1,3} \\
}
\affiliations {
    \textsuperscript{\rm 1}National University of Singapore,  \\
    \textsuperscript{\rm 2}Tencent Game AI Research Center,  \\
    \textsuperscript{\rm 3}Yale-NUS College \\
    e0321276@u.nus.edu, \{bohawkwang,brandonyang\}@tencent.com, robby.tan@nus.edu.sg
}

\begin{document}
% \linenumbers
\maketitle

\begin{abstract}
Despite the recent progress, 3D multi-person pose estimation from monocular videos is still challenging due to the commonly encountered  problem of missing information caused by occlusion, partially out-of-frame target persons, and inaccurate person detection.
To tackle this problem, we propose a novel framework integrating graph convolutional networks (GCNs) and temporal convolutional networks (TCNs) to robustly estimate camera-centric multi-person 3D poses that does not require camera parameters. 
In particular, we introduce a human-joint GCN, which unlike the existing GCN, is based on a directed graph that employs the 2D pose estimator's confidence scores to improve the pose estimation results.
We also introduce a human-bone GCN, which models the bone connections and provides more information beyond human joints.
The two GCNs work together to estimate the spatial frame-wise 3D poses, and can make use of both visible joint and bone information in the target frame to estimate the occluded or missing human-part information. 
To further refine the 3D pose estimation, we use our temporal convolutional networks (TCNs) to enforce the temporal and human-dynamics constraints.
We use a joint-TCN to estimate person-centric 3D poses across frames, and propose a velocity-TCN to estimate the speed of 3D joints to ensure the consistency of the 3D pose estimation in consecutive frames.
Finally, to estimate the 3D human poses for multiple persons, we propose a root-TCN that estimates camera-centric 3D poses without requiring camera parameters.
Quantitative and qualitative evaluations demonstrate the effectiveness of the proposed method.
Our code and models are available at \url{https://github.com/3dpose/GnTCN}. 
\end{abstract}

\section{Introduction}

% Top-down and bottom-up methods, and our focus is top-down approaches.
Significant progress has been made in 3D human pose estimation in recent years, e.g.  \cite{sun2019hrnet,pavllo20193d,cheng2019occlusion,cheng2020sptaiotemporal}. In general, existing methods can be classified as either top-down or bottom-up. Top-down approaches use human detection to obtain the bounding box of each person, and then perform pose estimation for every person. Bottom-up approaches are human-detection free and can estimate the poses of all persons simultaneously. Top-down approaches generally demonstrate more superior performance in pose estimation accuracy, and are suitable for many applications that require high pose estimation precision~\cite{pavllo20193d,cheng2020sptaiotemporal}; while bottom-up approaches are better in efficiency~\cite{cao2017realtime,cao2019openpose}. In this paper, we aim to further improve 3D pose estimation accuracy,  and thus push forward the frontier of the top-down approaches. 

\begin{figure}[t]
	\centering
	\includegraphics[width=8.5cm]{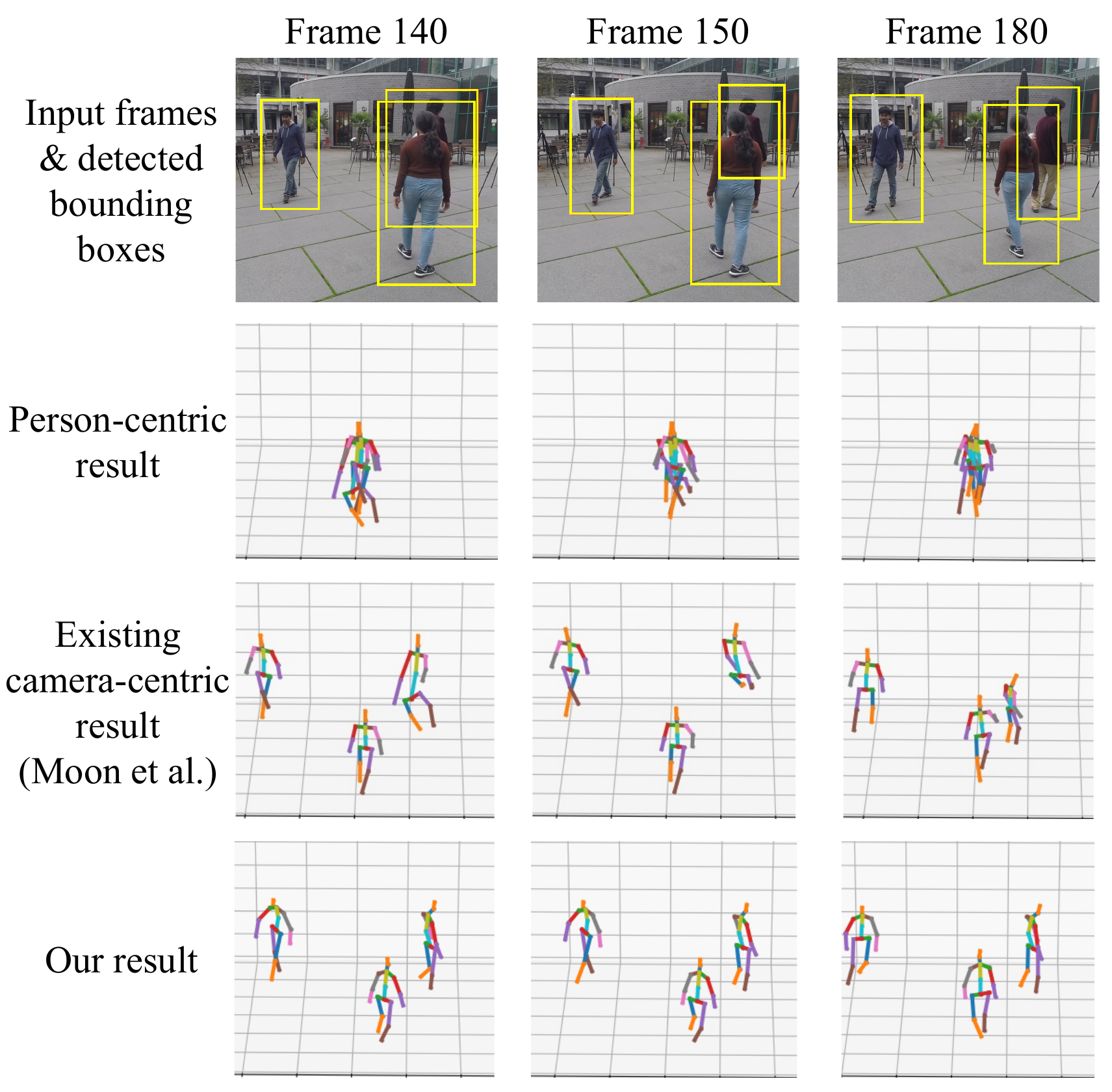}
	\caption{Incorrect 3D multi-person pose estimation caused by person-centric pose estimation or occlusions. The person-centric estimation loses the location of each person in the scene ($2^{nd}$ row) and existing method suffers from missing information due to occlusion ($3^{rd}$ row). }
	\label{fig:intriguing_example}
\end{figure}

Most top-down methods focus on single person and define a 3D pose in a person-centric coordinate system (e.g., pelvis-based origin), which cannot be extended to multiple persons. Since for multiple persons, all the estimated skeletons need to reside in a single common 3D space in correct locations. 
The major problem here is that by  applying the person-centric coordinate system, we lose the location of each person in the scene, and thus we do not know where to put them,  as shown in  Fig.~\ref{fig:intriguing_example}, second row.
Another major problem of multiple persons is the missing information of the target persons, due to occlusion, partially out-of-frame, inaccurate person detection, etc. 
For instance, inter-person occlusion may confuse human detection~\cite{lin2020hdnet,sarandi2020metrabs}, causing erroneous pose estimation~\cite{li2019crowdpose,umer2020self}, and  thus affect the 3D pose estimation accuracy (as shown in Fig.~\ref{fig:intriguing_example}, third row). Addressing these problems is critical for multi-person 3D pose estimation from monocular videos.

% Introduction of GCN module (joint and bone)
In this paper, we exploit the use of the visible human joints and bone information spatially and temporally utilizing GCNs (Graph Convolutional Networks) and TCNs (Temporal Convolutional Networks).  
Unlike most existing GCNs, which are based on undirected graphs and only consider the connection of joints, we introduce a directed graph that can capture the information of both joints and bones, so that the more reliably estimated joints/bones can influence the unreliable ones caused by occlusions (instead of treating them equally as in undirected graphs).
Our human-joint GCN  (in short, joint-GCN) employs the 2D pose estimator's heatmap confidence scores as the weights to construct the graph's edges, allowing the high-confidence joints to correct low-confidence joints in our 3D pose estimation.
While our human-bone GCN (in short, bone-GCN) makes use of the confidence scores of the part affinity field~\cite{cao2019openpose} to provide complementary information to the joint GCN. 
The features produced by the joint- and bone-GCNs are concatenated and fed into our fully connected layers to estimate a person-centric 3D human pose. 

Our GCNs focus on recovering the spatial information of target persons in a frame-by-frame basis. 
% The GCN can complete the human skeleton with adjacency based on human kinematics information.
%
To increase the accuracy across the input video, we need to put more constraints temporally, both in terms of the smoothness of the motions and the correctness of the whole body dynamics (i.e., human dynamics). 
To achieve this, we first employ a joint-TCN that takes a sequence of the 3D poses   produced by the GCN module as input, and estimate the person-centric 3D pose of the central frame. The joint-TCN imposes a smoothness constraint in its prediction and also imposes the constraints of human dynamics.
However, the joint-TCN can estimate only person-centric 3D poses (not camera-centric). Also, the joint-TCN is not robust to occlusion. 
To resolve the problems, we introduce two new types of TCNs: root-TCN and velocity-TCN.

Relying on the output of the joint-TCN, our root-TCN  produces the camera-centric 3D poses, where the  $X,Y,Z$ coordinates of the person center, i.e. the pelvis, are in the  camera coordinate system. The root-TCN is based on the weak perspective camera model, and does not need to be trained with large variation of camera parameters, since it estimates only the relative depth, $Z/f$.
Our velocity-TCN takes the person-centric 3D poses and the velocity from previous frames as input, and estimates the velocity at the current frame. 
Our velocity-TCN estimates the current pose based on the previous frames using motion cues. Hence, it is more robust to missing information, such as in the case of occlusion.
The reason is because the joint-TCN focuses on the correlations between past and future frames regardless of the trajectory, while the velocity-TCN focuses on the motion prediction, and thus makes the estimation more robust.

In summary, our contributions are listed as follows.
\begin{itemize}
	\item Novel directed graph-based joint- and bone-GCNs to estimate 3D poses  that can predict human 3D poses even though the information of the target person is incomplete due to occlusion, partially out-of-frame, inaccurate human detection, etc. 
	\item Root-TCN that can estimate the camera-centric 3D poses using the weak perspective projection without requiring camera parameters.
	\item Combination of velocity- and joint-TCNs that utilize velocity and human dynamics for robust 3D pose estimation.
\end{itemize}

\section{Related Works}

\paragraph{3D human pose estimation in video} 
Recent 3D human pose estimation methods utilize temporal information via recurrent neural network (RNN)~\cite{hossain2018exploiting,lee2018propagating,chiu2019action} or TCN~\cite{pavllo20193d,cheng2019occlusion,sun2019human,cheng2020sptaiotemporal} improve the temporal consistency and show promising results on single-person video datasets such as HumanEva-I, Human3.6M, and MPI-INF-3DHP~\cite{sigal2010humaneva,h36m_pami,mehta2017monocular}, but they still suffer from the inter-person occlusion issue when applying to multi-person videos. Although a few works take occlusion into account~\cite{ghiasi2014cvpr,charles2016cvpr,belagiannis2017fg,cheng2019occlusion,cheng2020sptaiotemporal}, in a top-down framework, it is difficult to reliably estimation 3D multi-person human poses in videos due to erroneous detection and occlusions. Moreover, none of these method estimate camera-centric 3D human poses.

\paragraph{Monocular 3D  human pose estimation}
Earlier approaches that tackle camera-centric 3D human pose from monocular camera require camera parameters as input or assume fixed camera pose to project the 2D posture into camera-centric coordinate~\cite{mehta2017vnect,mehta2019xnect,pavllo20193d}. As a result, these methods are inapplicable for wild videos where camera parameters are not available. Removing the requirement of camera parameters has drawn researcher's attention recently. Moon et al.~\cite{Moon_2019_ICCV_3DMPPE} first propose to learn a correction factor for a person's root depth estimation from a single image. Several recent works~\cite{li2020hmor,lin2020hdnet,zhen2020smap} show improved performance compared with~\cite{Moon_2019_ICCV_3DMPPE}. Li et al.~\cite{li2020hmor} develop an integrated method for detection, person-centric pose, and depth estimation from a single image. Lin et al.~\cite{lin2020hdnet} propose to formulate the depth regression as a bin index estimation problem. Zhen et al.~\cite{zhen2020smap} propose to estimate 2.5D representation of body parts first and then reconstruct 3D human pose. Unlike their approach, our method is video-based where temporal information is utilized by TCN on top of GCN output, which leads to improved 3D pose estimation. 

\begin{figure*}[ht]
    \centering
    \includegraphics[width=18cm]{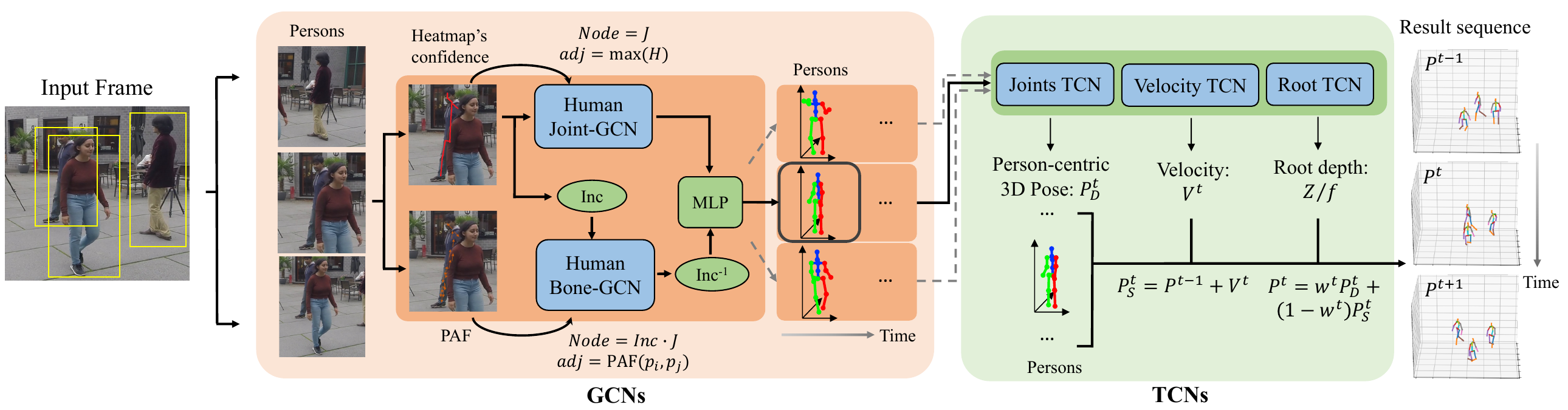}
    \caption{The framework of our approach. The 2D poses and part affinity field for each bounding box are fed into our joint- and bone-GCNs to obtain the full 3D poses (left). After obtaining all poses in the video, they are grouped by IDs which is provided by pose tracker, and fed into the the joint-, root- and velocity-TCN to obtain the camera-centric 3D pose estimation (right). }
    \label{fig:structure}
\end{figure*}

\paragraph{GCN for pose estimation}
Graph convolutional network (GCN) has been applied to 2D or 3D human pose estimation in recent years~\cite{zhao2019semantic,cai2019exploiting,ci2019optimizing,qiu2020dgcn}. Zhao et al.~\cite{zhao2019semantic} propose a graph neural network architecture to capture local and global node relationship and apply the proposed GCN for single-person 3D pose estimation from image. Ci et al~\cite{ci2019optimizing} explore different network structures by comparing fully connected network and GCN and develop a locally connected network to improve the representation capability for single-person 3D human pose estimation from image as well. 
Cai et al.~\cite{cai2019exploiting} construct an undirected graph to model the spatial-temporal dependencies between different joints for single-person 3D pose estimation from video data. Qiu et al.~\cite{qiu2020dgcn} develop a dynamic GCN framework for multi-person 2D pose estimation from a image. Our method is different from all these methods in terms of we propose to use directed graph to incorporate heatmap and part affinity field confidence in graph construction, which brings the benefit of overcoming the limitation of human detection on top-down pose estimation methods. 

\section{Method}
The overview of our framework is shown as Fig. \ref{fig:structure}. 
Having obtained the 2D poses from the 2D pose estimator, the poses are normalized so that they are centered at the root point, which is at the hip of human body. 
Each pose is then fed into our joint- and bone-GCNs to obtain its 3D full pose, despite the input 2D pose might be incomplete. 
Finally, a 3D full pose sequence is fed into the joint-TCN, root-TCN, and velocity-TCN to obtain the camera-centric 3D human poses that have smooth motion and comply with natural human dynamics. 

\subsection{Joint-GCN and Bone GCN}
Existing top-down methods are erroneous when  the target human bounding box is incorrect, due to missing information (occlusion, partially out-of-frame, blur, etc.). To address this common problem, we introduce joint-GCN and bone-GCN that can correct the 3D poses from inaccurate 2D pose estimator. These GCNs work on a frame-by-frame basis.

% \yb{I suggest add a figure to show the definitions and examples of graph and weights.}

Following the structure of the human body, we assign the coordinates $(x_i, y_i)$ of the human joints from the 2D pose estimator to each vertex of our graph, and establish connections between each pair of the joints.
Unlike most GCNs, which are based on an undirected graph,  we propose a GCN based on a directed graph.
The directed graph allows us to propagate information more from high-confident joints to low-confident ones, and thus reduces the risk of propagating erroneous information (e.g., occluded joints or missing joints) in the graph.
In other words, the low-confident joints contribute less to the message propagation than the high-confident ones. 
Details of the directed graph are available in the supplementary material. 

The joint-GCN uses the 2D joints as the vertices and the confidence scores of the 2D joints as the edge weights, while the bone-GCN uses the confidence scores of part affinity field \cite{cao2017realtime}) as the edge weights. 
The features produced by the two GCNs  are concatenated together and fed to a Multi Layer Perceptron to obtain the person-centric 3D pose estimation.

In GCNs, the message is propagated according to adjacent matrix, which indicates the edge between each pair of vertices. The adjacency matrix is formed by the following rule:

\begin{equation}
\mathbb{A}_{i,j} = 
\begin{cases}
max(H_i) e^{-order(i,j)} (i\ne j)\\
max(H_i) (i=j)\\
\end{cases},
\label{eq:adj}
\end{equation}

\noindent where $H$ is the heatmap from the 2D pose estimator. $order(i,j)$ stands for the number of the order of neighboring vertices, which means the number of hops required to reach vertex $j$ from vertex $i$. 
% The structure follows typical human kinematic structure where the pelvis is the root and limbs are the leaves, details are available in the supplementary material. 
This formation of adjacency imposes more weight for close vertices and less for distance vertices. 

The forward propagation of each GCN layer can be expressed as:

\begin{equation}
h_{i} = \sigma (F(h_{i-1}) W_{i}),
\end{equation}

\noindent where $F$ is the feature transformation function, and $W$ is the learnable parameter of layer $i$. To learn a network with strong generalization ability, we follow the idea of Graph SAGE \cite{hamilton2017inductive} to learn a generalizable aggregator, which is formulated as:

\begin{equation}
F(h_{i}) = \widetilde{\mathbb{A}} h_{i} \oplus h_{i},
\end{equation}

\noindent where $h_i$ is the output of layer $i$ in the GCN and $\oplus$ stands for the concatenation operation. $\widetilde{\mathbb{A}}$ is the normalized adjacency matrix. Since our  method is based on a directed graph, which uses a non-symmetric adjacency matrix, the normalization is $\widetilde{\mathbb{A}}_{i,j}=\frac{\mathbb{A}_{i,j}}{D_{j}}$ instead of
 $\widetilde{\mathbb{A}}_{i,j}=\frac{\mathbb{A}_{i,j}}{\sqrt{D_{i}D_{j}}}$ in \cite{kipf2016semi}, $D_i$ and $D_j$ are the indegree of vertices $i$ and $j$, respectively. 
This normalization ensures that the indegree of each vertex sums to $1$, which prevents numerical instability. 

Our joint-GCN considers only human-joints and does not include the information of bones, which can be critical for the cases when the joints are missing due to occlusion or other reasons. To exploit the bone information, we created a bone-GCN.
First, we construct the incidence matrix $\mathbb{I}_n$ of shape $[\#bones, \#joints]$ to represent the bone connections, where each row represents an edge and the columns represent vertices. 
For each bone, the parent joint is assigned with $-1$ and the child joint is assigned with $1$. 
Second, the incidence matrix $\mathbb{I}_n$ is multiplied with the joint matrix $J$ to obtain the bone matrix $\mathbb{B}$, which will be further fed into our bone-GCN. 

In joint matrix $\mathbb{J}$, each row stands for the 2D coordinate $(x,y)$ of a joint. 
Unlike our joint-GCN, where the adjacency matrix is drawn from the joint heatmap produced by 2D pose estimator, our human-bone GCN utilizes the confidence scores from the part affinity field, following the method of \cite{cao2017realtime}, as the adjacency. 
Finally,  the outputs from our human-joint GCN and human-bone GCN are concatenated together and fed into an MLP (Multi-layer Perceptron).
The loss function we use is the $L2$ loss between GCN 3D joints output $P_{GCN}$ and 3D ground-truth skeleton $\widetilde{P}$, which is $L_{GCN} = ||\widetilde{P} - P_{GCN}||_2^2.$

In the training stage, to obtain sufficient variation and to increase the robustness of our GCNs, we use not only the results from our 2D pose estimator, but also augmented data from our ground-truths. Each joint is assigned with a random confidence score and random noise.

\subsection{Root-TCN}
In most of the videos, the projection can be modelled as weak perspective:

\begin{equation}
\left[ \begin{array}{c} x \\ y \\ 1 \end{array} \right]=1/Z \begin{bmatrix} f & 0 & c_x \\ 0 & f & c_y \\ 0 & 0 & 1 \end{bmatrix} \left[ \begin{array}{c} X \\ Y \\ Z \end{array}\right],
\label{eq:proj}
\end{equation}

\noindent where $x$ and $y$ are the image coordinates, $X,Y$ and $Z$ are the camera coordinates. $f, c_x, c_y$ stands for the focal length and camera centers, respectively. Thus we have:

\begin{equation}
\label{equ:world}
X = \frac{Z}{f}(x-c_{x}) \quad
Y = \frac{Z}{f}(y-c_{y}).
\end{equation}

By assuming $(c_x, c_y)$ as the image center, which is applicable for most cameras, the only parameters we need to estimate is depth $Z$ and focal length $f$. 
To be more practical, we jointly estimate $Z/f$, instead of estimating them separately.
This enables our method to be able to take wild videos that the camera parameters are unknown.

According to the weak perspective assumption, the scale of a person in a frame indicates the depth in the camera coordinates. 
Hence, we propose a network, root Temporal Convolutional Network (root-TCN), to estimate the $Z/f$ from 2D pose sequences. 
We first normalize each 2D pose by scaling the average joint-to-pelvis distance to $1$, using a scale factor $s$. 
Then we concatenate the normalized pose $p$, scale factor $s$, as well as the person's center in the frame as $c$, and feed a list of such concatenated features in a local temporal window into the TCN for depth estimation in the camera coordinates.

As directly learning $Z/f$ is not easy to converge, we transform this regression problem into a classification problem. 
For each video, we divide the depth into $N$ discrete ranges, set to 60 in our experiments, and our root-TCN  outputs a vector with length $N$ as $\{x_1, ..., x_N\}$, where $x_i$ indicates the probability that $Z/f$ is within the $i$th discrete range. 
Then, we apply \textit{Soft-argmax} to this vector to get the final continuous estimation of the depth as:

\begin{equation}
[\frac{Z}{f}]^t = \mathrm{\textit{Soft}\mbox{-}\textit{argmax}}\thinspace (\thinspace f_R(\thinspace p^{t-n:t+n}, c^{t-n:t+n}, s^{t-n:t+n})),
\end{equation}

\noindent where $t$ is the time stamp, and $n$ is half of the temporal window's size. 
This improves the training stability and reduces the risk of large errors. 

The loss function for the depth estimation is defined as the mean squared error between the ground truth and predictions, expressed as $L_{Root} = (\frac{Z}{f}-\frac{\hat{Z}}{\hat{f}})^2$, where $Z/f$ is the predicted value, and $\hat{Z}/\hat{f}$ denotes the ground truth. According to Eq.(\ref{equ:world}), we can calculate the coordinates for the person's center as $P^t_D$.

\subsection{Joint-TCN and Velocity-TCN}
To increase the accuracy of the 3D poses across the input video, we impose temporal constraints, by employing a temporal convolutional network (TCN) \cite{cheng2020sptaiotemporal} that takes a sequence of consecutive 3D poses as input. 
We call this TCN a joint-TCN, which is trained using various 3D poses and their augmentation, and hence capture human dynamics. The joint-TCN outputs the person-centric 3D pose, $P_D$. The TCN utilizes temporal information to interpolate the poses of occluded frames with temporal information.

However, when persons get close and occlude each other, there may be fewer visible joints belonging to a person and more distracting joints from other persons.
To resolve the problem, in addition to the joint-TCN, we propose a velocity-based estimation network, velocity-TCN, which takes the 3D joints and their velocities as input, and predicts the velocity of all joints as:

\begin{equation}
V^t = (v_x^t, v_y^t, v_z^t) = \mbox{TCN}_v(p^{t-n:t-1}, V^{t-n:t-1}),
\end{equation}

\noindent where $p$ stands for the 2D pose and $V^t$ denotes the velocity at time $t$. TCN$_v$ is the velocity-TCN. The velocity here is proportional to $1/f$ according to Eq. (\ref{equ:world}). We normalize the velocity both in training and testing. 
With estimated $V^t$, we can obtain the coordinate $P^t_S=P^{t-1}+V^t$, where $P^t_S$ and $P^{t-1}$ are estimated coordinates at time $t$ and $t-1$. The calculation of $P^{t-1}$ is discussed  later in Eq.(\ref{equ:root}).

The joint-TCN predicts the joints by interpolating the past and future poses, while our velocity-TCN predicts the future poses using motion cues. Both of them are able to handle the occlusion frames, but the joint-TCN focuses on the connection between past and future frames regardless of the trajectory, while the velocity-TCN focuses on the motion prediction, which can handle a motion drift.
To leverage the benefits of both, we introduce an adaptive weighted average of their estimated  coordinates $P^t_D$ and $P^t_S$.  

We utilize the 2D pose tracker \cite{umer2020self} to detect and track human poses in the video. 
In the tracking procedure, we regard the heatmaps with less than $0.5$
confidence value as occluded joints, and the pose with less than $30\%$ non-occluded joints as the occluded pose. By doing this, we obtain the occlusion information for both joints and frames. Note that the values here are obtained empirically through our experiments.
Suppose we find an occlusion duration from $T^{start}_{occ}$ to $T^{end}_{occ}$ for a person, then we generate the final coordinates as:

\begin{equation}
\label{equ:root}
P^t = w^t P^t_D + (1 - w^t) P^t_S,
\end{equation}

\noindent where $w^t_v = e^{-min(t-T^{start}_{occ},T^{end}_{occ}-t)}$. For frames that are closer to occlusion duration boundaries, we trust $P^t_D$ more; for those are far from occlusion boundaries, we trust $P^t_S$ more.
The velocity-TCN loss is the $L2$ loss between the predicted 3D points at $t$ and ground-truth 3D points.

\section{Experiments}

\textbf{MuPoTS-3D} is a 3D multi-person testing set with both indoor and outdoor scenes~\cite{mehta2018single}. 
% and it's corresponding training set is augmented from 3DHP, called MuCo-3DHP. 
The ground-truth 3D pose of each person in the video is obtained from multi-view markerless capture, which is suitable for evaluating 3D multi-person pose estimation performance in both person-centric and camera-centric coordinates. Unlike previous methods~\cite{Moon_2019_ICCV_3DMPPE} using the training set (MuCo-3DHP) to train their models and then do evaluation on MuPoTS-3D, we use MuPoTS-3D for testing only without fine-tuning.

\vspace{0.2em}
\noindent \textbf{3DPW} is an outdoor multi-person dataset for 3D human pose reconstruction~\cite{3DPW}. 
% that contains 24,12,24 video clips for training, validation, and testing. The 3D ground-truth is obtained by IMU readings from a person performs daily activities outdoor with the sensor in each video clip, so it is suitable for in-the-wild video evaluation. 
Following previous methods~\cite{humanMotionKanazawa19,sun2019human}, we use 3DPW for testing only without any fine-tuning. 
The ground-truth of 3DPW is SMPL 3D mesh model~\cite{loper2015smpl}, where the definition of joints differs from the one commonly used in 3D human pose estimation (skeleton-based) like Human3.6M~\cite{tripathi2020posenet3d}, so it is unfair to evaluate skeleton-based methods on it even after joint adaption or scaling. 
To perform a fair comparison, we select an occlusion subset from the 3DPW test set (please refer to the supplementary material for details). And the performance change of a method between the full test set and the subset indicates how well the method can handle the missing information problem caused by occlusions.
% As we focus on solving the missing information problem caused by occlusions and incorrect detections, but evaluation value on this subset is still not a good performance indicator, the performance change of a method between full testing set and the subset can measure how well the method can handle the missing information problem.

%Human3.6M dataset.
\noindent \textbf{Human3.6M} is a widely used dataset and benchmark for 3D human pose estimation~\cite{h36m_pami}. It contains 3.6 million single-person indoor images captured by the MoCap system, which is suitable for evaluation of single-person pose estimation and camera-centric coordinates prediction. 
% As our method focuses on 3D multi-person problem in camera-centric coordinates, Human3.6M is used mostly for evaluating the performance of camera-centric coordinate prediction. 
Following previous works~\cite{hossain2018exploiting,pavllo20193d,wandt2019repnet}, the subject 1,5,6,7,8 are used for training, and 9 and 11 for testing.

\vspace{0.2em}
% Please refer to the supplementary material for person-centric and camera-centric evaluation metrics used.
\noindent \textbf{Evaluation and Implementation} MPJPE, PA-MPJPE, PCK, and $AUC_{rel}$ are used for person-centric pose estimation evaluation. $AP_{25}^{root}$ and PCK$_{abs}$ are used for camera-centric pose estimation evaluation. 
% used include MPJPE, PA-MPJPE, PCK, $AUC_{rel}$, $AP_{25}^{root}$, and PCK$_{abs}$. 
Each GCN and TCN is trained for 100 epochs with initial learning rate $1e-3$, more details are available in the supplementary material. 

\begin{table}
	\footnotesize
	\centering
	\begin{tabular}{c|c|c|c|c}
		\cline{1-5}
		\rule{0pt}{2.6ex}
		\textbf{Method} & $AP_{25}^{root}$ & $AUC_{rel}$ & PCK & PCK$_{abs}$\\
		\cline{1-5}
		\rule{0pt}{2.6ex}
		Baseline & 24.1 & 32.9 & 74.4 & 29.8\\
		Baseline (GT box) & 28.5 & 34.2 & 78.9 & 31.2\\
		Baseline + GCNs  & 35.4 & 39.7 & 83.2 & 35.1\\
		Baseline + TCNs  & \underline{38.4} & \underline{43.1} & \underline{85.3} & \underline{38.7}\\
		Full model & \textbf{45.2} & \textbf{48.9} & \textbf{87.5} & \textbf{45.7}\\
		\cline{1-5}
		% GT & \textbf{100} & \textbf{47.0} & \textbf{87.2} & \textbf{87.2}\\
	\end{tabular}
	\caption{Ablation study on MuPoTS-3D dataset. Best in bold, second best underlined.}
	\label{tab:Ablation}
\end{table}

\vspace{0.2em}
\noindent \textbf{Ablation Studies}
In Table~\ref{tab:Ablation}, we provide the results of an ablation study to validate the major components of the proposed framework. 
MuPoTS-3D is used as it has been used for 3D multi-person pose evaluation in person-centric and camera-centric coordinates~\cite{Moon_2019_ICCV_3DMPPE}. $AUC_{rel}$ and PCK metrics are used to evaluate person-centric 3D pose estimation performance, $AP_{25}^{root}$ and PCK$_{abs}$ metrics are used to evaluate camera-centric 3D pose (i.e., camera-centric coordinate) estimation following~\cite{Moon_2019_ICCV_3DMPPE}.

\begin{table}
	\footnotesize
	\centering
	\begin{tabular}{c|c|c|c|c}
		\cline{1-5}
		\rule{0pt}{2.6ex}
		\textbf{Method} & $AP_{25}^{root}$ & $AUC_{rel}$ & PCK & PCK$_{abs}$\\
		\cline{1-5}
		\rule{0pt}{2.6ex}
		Joint* GCN & 24.1 & 27.3 & 73.1 & 25.6\\
	    Joint  GCN & 28.5 & 30.1 & 76.8 & 29.0\\
	    Joint + Bone* GCN & 28.4 & 31.9 & 78.1 & 29.7 \\
		Joint + Bone  GCN & 33.4 & 37.9 & 82.6 & 34.3\\
		Joint + Bone + Aug. & 35.4 & 39.7 & 83.2 & 35.1\\  \hline
		Joint TCN & \underline{43.1} & \underline{45.8} & \underline{86.2} & \underline{42.6} \\
		Joint + Velocity & \textbf{45.2} & \textbf{48.9} & \textbf{87.5} & \textbf{45.7}\\
		\cline{1-5}
		% GT & \textbf{100} & \textbf{47.0} & \textbf{87.2} & \textbf{87.2}\\
	\end{tabular}
	\caption{Ablation study on our proposed Joint and Bone GCNs and TCNs. * stands for the GCN structure with undirected graph. We keep the GCN as the best one (joint + bone + aug.) to perform an ablation study on TCN.}
	\label{tab:Ablation_GCN_TCN}
\end{table}

In particular, we use the joint-TCN with time window 1 plus a root-TCN with time window 1 as a baseline for both person-centric and camera-centric coordinate estimation as shown at the 1st row in Table~\ref{tab:Ablation}. We use the baseline with ground-truth bounding box (i.e., perfect 2D tracking) as a second baseline to illustrate even with perfect detection bounding box the baseline still performs poorly because it cannot deal with occlusion and distracting joints from other persons. On the contrary, we can see significant performance (e.g., $18\% \sim 29\%$ improvement against the baseline in PCK$_{abs}$) improvements after adding the proposed GCN and TCN modules as shown in row 3 - 5 in Table~\ref{tab:Ablation}. The benefits from the TCNs are slightly larger than those from the joint and bone GCNs as temporal information is used by the TCNs while GCNs only use frame-wise information. Lastly, our full model shows the best performance, with $53\%$ improvement against the baseline in terms of PCK$_{abs}$.

We perform a second ablation study to break down the different pieces in our GCN and TCN modules to show the effectiveness of each individual component in Table \ref{tab:Ablation_GCN_TCN}. We observe the undirected graph-based GCN performs the worst as shown in the 1st row. Our joint-GCN, bone-GCN, and data augmentation (applying random cropping and scaling) show improved performance in row 2 - 4. On top of the full GCN module, we also show the contribution of the joint-TCN and velocity-TCN in row 5 - 6 (row 6 is our full model). Similar to Table \ref{tab:Ablation} we can see the TCN module brings more improvement compared with the GCN module as temporal information is used.

\begin{table}[t]
	\footnotesize
	\centering
	\begin{tabular}{c|c|c|c}
		\cline{1-4}
		\cline{1-4}
		\rule{0pt}{2.6ex}
		\textbf{Group} & \textbf{Method} & PCK & PCK$_{abs}$\\
		\cline{1-4}
		\rule{0pt}{2.6ex}
		% Rogez et al., CVPR~\shortcite{rogez2017lcr} & N/A & 53.8 & N/A\\
		& Mehta et al. \shortcite{mehta2018single} & 65.0 & n/a\\
		Person- & Rogez et al., \shortcite{rogez2019lcr} & 70.6 & n/a\\
		% Mehta et al. arXiv~\shortcite{mehta2019xnect} & N/A & 70.4 & N/A\\
		centric & Cheng et al. \shortcite{cheng2019occlusion} & 74.6 & n/a\\
		& Cheng et al. \shortcite{cheng2020sptaiotemporal} & 80.5 & n/a\\
		\cline{1-4}
		& Moon et al. \shortcite{Moon_2019_ICCV_3DMPPE} & 82.5 & 31.8\\
		
		Camera- & Lin et al. \shortcite{lin2020hdnet} & \underline{83.7} & 35.2\\
		centric & Zhen et al. \shortcite{zhen2020smap} & 80.5 & 38.7\\
		& Li et al. \shortcite{li2020hmor} & 82.0 & \underline{43.8}\\
		& Our method & \textbf{87.5} & \textbf{45.7}\\
		\cline{1-4}
		\cline{1-4}
		% GT & \textbf{100} & \textbf{47.0} & \textbf{87.2} & \textbf{87.2}\\
	\end{tabular}
	\caption{Quantitative evaluation on multi-person 3D dataset, MuPoTS-3D. Best in bold, second best underlined.}
	\label{tab:MuPoTS_3d}
\end{table}

\vspace{0.2em}
\noindent \textbf{Quantitative Results}
To compare with the state-of-the-art (SOTA) methods in 3D multi-person human pose estimation, we perform evaluations on MuPoTS-3D as shown in Table~\ref{tab:MuPoTS_3d}. Please note our network is trained only on Human3.6M to have a fair comparison with other methods~\cite{cheng2019occlusion,cheng2020sptaiotemporal}. As the definition of keypoints in MuPoTS-3D is different from the one where our model is trained on, we use joint adaptation \cite{tripathi2020posenet3d} to transfer the definition of keypoints. Among the methods in Table~\ref{tab:MuPoTS_3d}, \cite{Moon_2019_ICCV_3DMPPE,lin2020hdnet,li2020hmor} are fine-tuned on 3D training set MuCo-3DHP. 

Regarding to the performance on MuPoTS-3D, our camera-centric pose estimation accuracy beat the SOTA \cite{li2020hmor} by $4.3\%$ on PCK$_{abs}$. A few papers reported their results on $AP_{25}^{root}$, where \cite{Moon_2019_ICCV_3DMPPE} is 31.0, \cite{lin2020hdnet} is 39.4, and our result is 45.2, where we beat the SOTA \cite{lin2020hdnet} by $14.7\%$. We also compare with other methods on person-centric 3D pose estimation, and get improvement of $4.5\%$ on PCK against the SOTA \cite{lin2020hdnet}. Please note we do not fine-tune on MuCo-3DHP like others~\cite{Moon_2019_ICCV_3DMPPE,lin2020hdnet,li2020hmor}, which is the training dataset for evaluation on MuPoTS-3D. Moreover, the SOTA method \cite{lin2020hdnet} on person-centric metric PCK shows poor performance on PCK$_{abs}$ (ours vs theirs: 45.7 - 35.2, $29.8\%$ improvement), and the SOTA method \cite{li2020hmor} on camera-centric metric PCK$_{abs}$ has mediocre performance on PCK (ours vs theirs: 87.5 vs 82.0, $6.7\%$ improvement). All of these results clearly show that our method not only surpasses all existing methods, but also is the only method that is well-balanced in both person-centric and camera-centric 3D multi-person pose estimation.

\begin{table}[t]
	\footnotesize
	\centering
	\begin{tabular}{c|c|c|c}
		\cline{1-4}
		\rule{0pt}{2.6ex}
		\textbf{Dataset} & \textbf{Method} & PA-MPJPE & $\delta$ \\
		\cline{1-4}
		\rule{0pt}{2.6ex}
		% & Bogo et al., ECCV'16~\cite{bogo2016keep}  & 106.8  \\
		% & Martinez et al., ICCV'17~\cite{martinez2017simple}  & 157.0  \\
		& Dabral et al. \shortcite{dabral2018learning}  & 92.2 & n/a \\
		& Doersch et al. \shortcite{doersch2019sim2real}  & 74.7 & n/a \\
		& Kanazawa et al. \shortcite{humanMotionKanazawa19} & 72.6 & n/a \\
		% & Arnab et al., CVPR'19~\cite{arnab2019exploiting} & 72.2 \\
		& Cheng et al. \shortcite{cheng2020sptaiotemporal} & 71.8 & n/a \\
		Original & Sun et al. \shortcite{sun2019human} & 69.5 & n/a \\
		& Kolotouros et al. \shortcite{kolotouros2019learning}* & \underline{59.2} & n/a \\
		& Kocabas et al., \shortcite{kocabas2020vibe}* & \textbf{51.9} & n/a \\
		& Our method  & 64.2 & n/a \\
		\cline{1-4}
		& Cheng et al. \shortcite{cheng2020sptaiotemporal} & 96.1 & \underline{+24.1}\\
		& Sun et al. \shortcite{sun2019human} & 94.1 & +24.6\\
		Subset & Kolotouros et al. \shortcite{kolotouros2019learning}*  & 88.9 & +29.7\\
		& Kocabas et al., \shortcite{kocabas2020vibe}* & \textbf{82.5} & +30.6 \\
		& Our method  & \underline{85.7} & \textbf{+21.5}\\
		\cline{1-4}
	\end{tabular}
	\caption{Quantitative evaluation using PA-MPJPE in millimeter on original 3DPW test set and its occlusion subset. * denotes extra 3D datasets were used in training. Best in bold, second best underlined.}
	\label{tab:3dpw}
\end{table}

3DPW dataset~\cite{3DPW} is a new 3D multi-person human pose dataset that contains multi-person outdoor scenes for person-centric pose estimation evaluation. Following previous works~\cite{humanMotionKanazawa19,sun2019human}, 3DPW is only used for testing and the PA-MPJPE values on test set are shown in Table~\ref{tab:3dpw}. As discussed in the Datasets section, the ground-truth definitions are different between 3D pose reconstruction and estimation where the ground-truth of 3DPW is SMPL mesh model, even we follow \cite{tripathi2020posenet3d} to perform joint adaptation to transform the estimated joints but still have a disadvantage, and the PA-MPJPE values cannot objectively reflect the performance of skeleton-based pose estimation methods. 

As aforementioned, we select a subset out of the original test set with the largest detection errors, and run the code of the top-performing methods in Table~\ref{tab:3dpw} on this subset for comparison. Table~\ref{tab:3dpw} shows that even with the disadvantage of different definition of joints, our method is the 3rd best on the original testing test, and becomes the 2nd best on the subset where the difference to the best one~\cite{kocabas2020vibe} is greatly shrunk. More importantly, the $\delta$ of PA-MPJPE between the original testing set and the subset in the 4th column in Table~\ref{tab:3dpw}, our method shows the least error increase compared with all other top-performing methods. In the particular, the two best methods~\cite{kolotouros2019learning,kocabas2020vibe} on the original testing set show 29.7 and 30.6 mm error increase while our method shows only 21.5 mm. The performance change of PA-MPJPE between the original testing set and the subset clearly demonstrates that our method is the best in terms of solving the missing information problem which is critical for 3D multi-person pose estimation.  

\begin{table}[h]
	\footnotesize
	\centering
	\begin{tabular}{c|c|c|c}
		\cline{1-4}
		\rule{0pt}{2.6ex}
		\textbf{Group} & \textbf{Method} & MPJPE & PA-MPJPE\\
		\cline{1-4}
		\rule{0pt}{2.6ex}
		% Pavlakos et al.~\cite{pavlakos2018ordinal} CVPR'18 & N/A & 56.2 & 41.8\\
		& Hossain et al., \shortcite{hossain2018exploiting} & 51.9 & 42.0 \\
		& Wandt et al., \shortcite{wandt2019repnet}* & 50.9 & 38.2 \\
		\small{Person-} & Pavllo et al., \shortcite{pavllo20193d} & 46.8 & 36.5 \\
		\small{centric} & Cheng et al., \shortcite{cheng2019occlusion} & 42.9 & 32.8\\
		& Kocabas et al., \shortcite{kocabas2020vibe} & 65.6 & 41.4  \\
		& Kolotouros et al. \shortcite{kolotouros2019learning} & \underline{41.1} & n/a \\
		\cline{1-4}
		& Moon et al., \shortcite{Moon_2019_ICCV_3DMPPE} & 54.4 & 35.2 \\
		\small{Camera-} & Zhen et al., \shortcite{zhen2020smap} & 54.1 & n/a \\
		\small{centric} & Li et al., \shortcite{li2020hmor} & 48.6 & \underline{30.5} \\
		& Ours & \textbf{40.9} & \textbf{30.4} \\
		\cline{1-4}
	\end{tabular}
	\caption{Quantitative evaluation on Human3.6M for normalized and camera-centric 3D human pose estimation. * denotes ground-truth 2D labels are used. Best in bold, second best underlined.}
	\label{tab:h36m}
\end{table}

\begin{figure*}[t]
	\centering
	\makebox[\textwidth]{\includegraphics[width=0.94\textwidth]{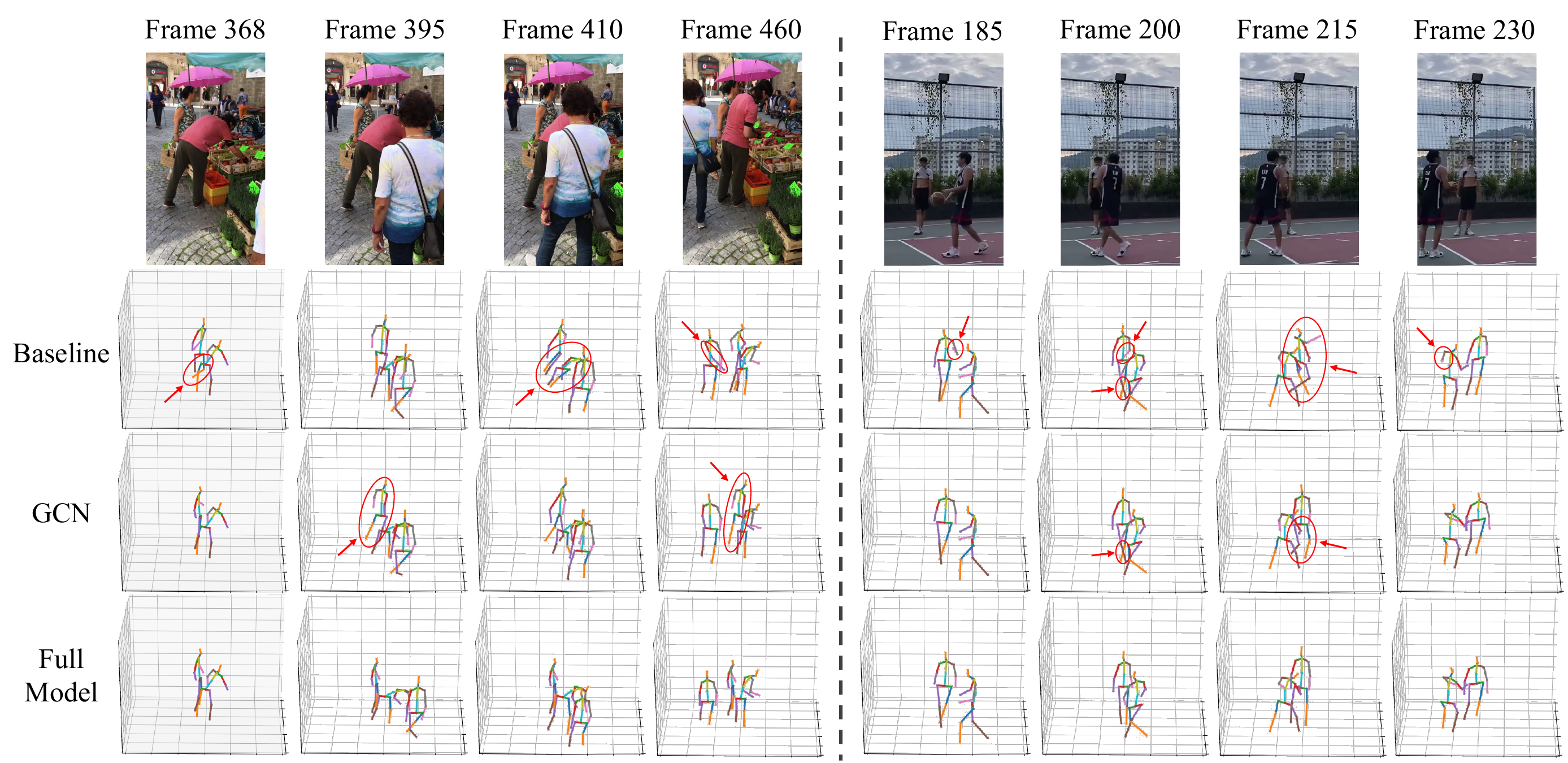}}
	\caption{Examples of results from our whole framework compared with different baseline results. First row shows the images from two video clips; second row shows the results from the baseline described in Ablation Studies; third row shows the result of the GCN module; last row shows the results of the whole framework. Wrong estimations are labeled with red circles.}
	\label{fig:qualitative_evaluation}
\end{figure*}

In order to further illustrate the effectiveness of both person-centric and camera-centric 3D pose estimation of our method, we perform evaluations on the widely used single-person dataset, Human3.6M. To evaluate camera-centric pose estimation, we use mean root position error (MPRE), a new evaluation metric proposed by \cite{Moon_2019_ICCV_3DMPPE}. Our result is 88.1 mm, the result of \cite{Moon_2019_ICCV_3DMPPE} is 120 mm, the result of \cite{lin2020hdnet} is 77.6 mm. Our method outperforms the result of \cite{Moon_2019_ICCV_3DMPPE} by a large margin: 31.9 mm error reduction and 26\% improvement. Although depth estimation focused method \cite{lin2020hdnet} shows better camera-centric performance on this single-person dataset Human3.6M, their camera-centric result on multi-person dataset MuPoTS-3D is much worse than ours (ours vs. theirs in PCK$_{abs}$: 45.7 - 35.2, $29.8\%$ improvement). Camera-centric 3D human pose estimation is for multi-person pose estimation, good performance only on single-person dataset is not enough to solve the problem.
% Please note Human3.6M as a single-person dataset cannot objectively show the strength of our method like multi-person 3D datasets MuPoTS-3D and 3DPW. our method is designed deal with occlusions and inaccurate detection, so 

To compare with most of the existing methods that evaluate person-centric 3D pose estimation on Human3.6M using MPJPE and PA-MPJPE, we report our results using the same metrics in Table~\ref{tab:h36m}. 
As Human3.6M contains only single-person videos, we do not expect to our method bring much improvement. It is observed that our method is comparable with the SOTA methods. 
% We also report our results on Human3.6M using the person-centric evaluation metrics (MPJPE and PA-MPJPE) in Table~\ref{tab:h36m} to compare with most of the existing methods. 
% Please note that the 2D datasets used in our training has slightly different joint definition compared to Human3.6M and the 2D detection is not fine-tuned on the Human3.6M, which leads to small higher error against the result of~\cite{cheng2020sptaiotemporal}. \textcolor{red}{May need some other references}
% Revision based on reviewers' comments
% R5.1 - adding kinematic constraints discussion
In addition, although our method shows improved performance over others that use kinematic constraints~\cite{wandt2019repnet,cheng2019occlusion} because of our GCNs and TCNs, adding kinematic constraints could potentially improve our performance further~\cite{akhter2015pose,kundu2020kinematic}. 
% Comparing with the methods using kinematic constraints~\cite{wandt2019repnet,cheng2019occlusion}, our method shows improved performance even without it because of our GCNs and TCNs. Adding kinematic constraints could potentially improve our performance further~\cite{akhter2015pose,kundu2020kinematic}. 

\vspace{0.2em}
\noindent \textbf{Qualitative Results}
As shown in Figure~\ref{fig:qualitative_evaluation}, our full model can better handle occlusions and incorrect detection compared with the baselines and the relative positions among all persons are well captured without camera parameters. More comparisons against SOTA methods and qualitative results on wild videos are available in the supplementary material.
% More results are shown in Figure~\ref{fig:qualitative_evaluation}. \textcolor{red}{Add more description here.}

\section{Conclusion}

We propose a new framework to unify  GCNs and TCNs for camera-centric 3D multi-person pose estimation. The proposed method successfully handles missing information due to occlusion, out-of-frame, inaccurate detections, etc.,  in videos and produces continuous pose sequences. Experiments on different datasets validate the effectiveness of our framework as well as our individual modules.

\section{Acknowledgements}
This research/project is supported by the National Research Foundation, Singapore under its Strategic Capability Research Centres Funding Initiative. Any opinions, findings and conclusions or recommendations expressed in this material are those of the author(s) and do not reflect the views of National Research Foundation, Singapore.

{\small
\bibliography{egbib}
}

\newpage

\subfile{supp.tex}

\end{document}

%% file: supp.tex
% \linenumbers
% \maketitle

\section{Supplementary Material}

In the following, additional information regarding implementation details, visualization of the occlusion subset, additional quantitative and qualitative evaluations, pose tracking illustration, and failure cases are provided as supplementary materials to better understand our method. 

\section{Implementation Details}
\label{sec:training_details}

\subsection{Graph Convolutional Layers}
\begin{figure}[h!]
    \centering
    \includegraphics[width=0.8\linewidth]{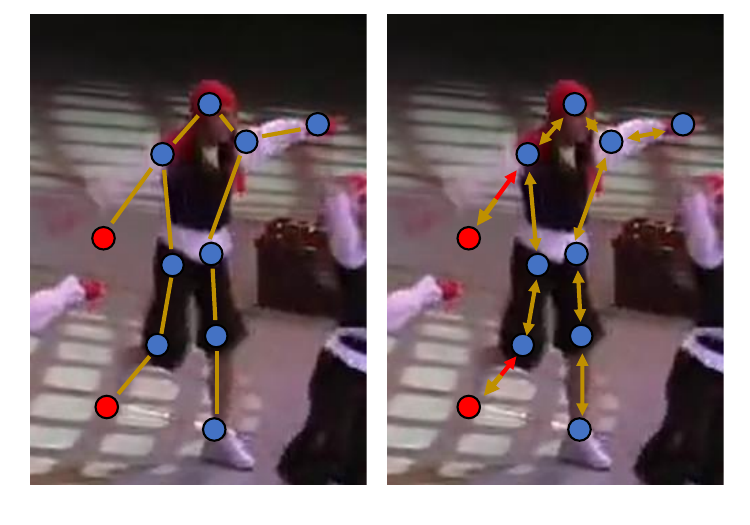}
    \caption{Difference between the undirected and directed graphs. The undirected graph propagates the message with same weight while the directed graph constrains the outward propagation for low-confident joints.}
    \label{fig:graph}
\end{figure}

For the Graph Convolutional Networks (GCNs), we utilize two branches which incorporate the confidence scores from heatmap and Part Affinity Field, respectively, named the joint-GCN and bone-GCN. Different from previous methods which use an undirected graph for the feature propagation, we propose to use  directed graphs in our GCNs. As shown in Fig. \ref{fig:graph}, the outward edges of the low-confident joints have the same weights as the high-confident joints in the conventional undirected graphs, even though the low-confident ones are probably wrongly estimated. In contrast, with the help of directed graphs, the outward edges of the low-confident joints have less weights, thus have less impact on the feature propagation. 

\begin{figure}[t]
    \centering
    \includegraphics[width=\linewidth]{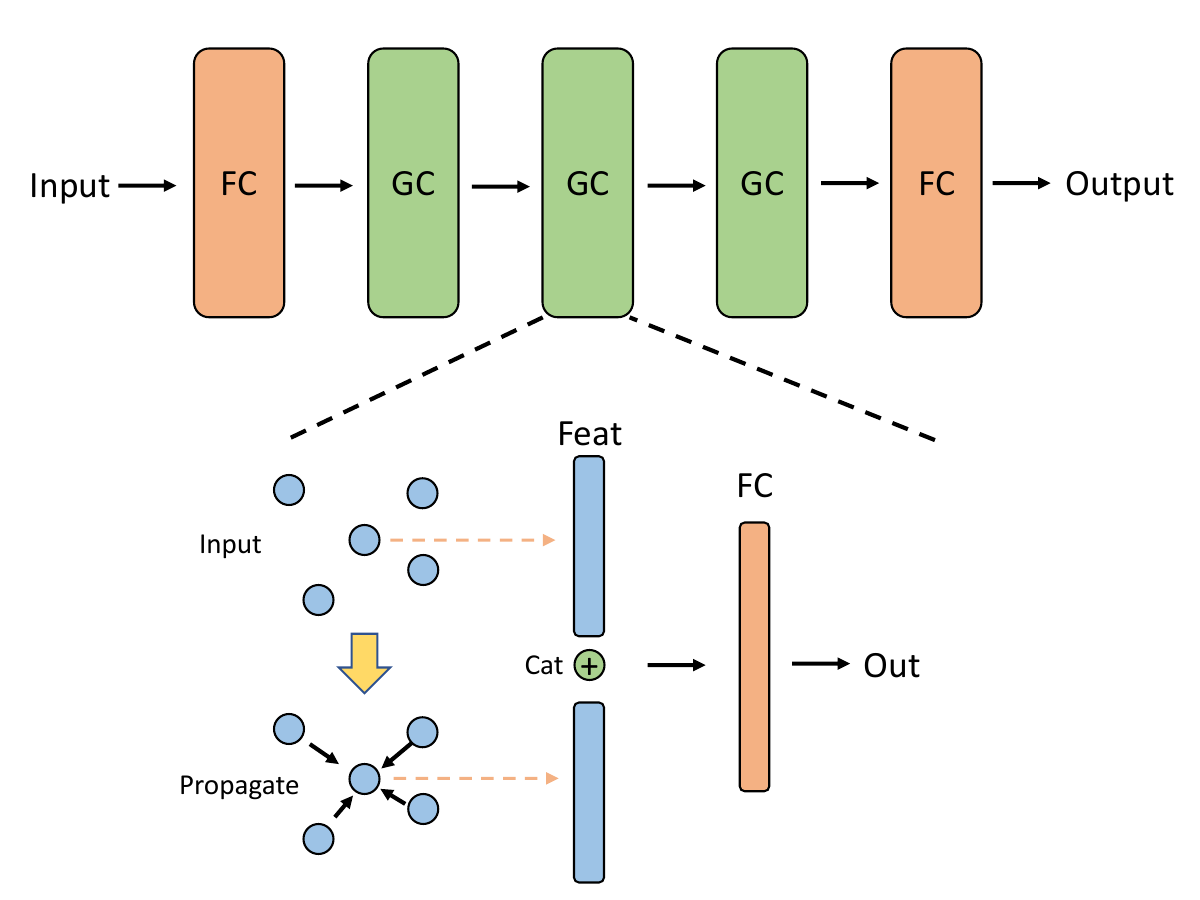}
    \caption{Structure of the GCN branch. FC represents the fully connected layer and GC represents the graph convolutional layer.}
    \label{fig:GCNbranch}
\end{figure}

The detailed structure of our GCN branch is shown in Fig. \ref{fig:GCNbranch}. The network consists of two encoding fully connected (FC) layers at both sides and several graph convolutional (GC) layers in between. Inspired by Graph SAGE \cite{hamilton2017inductive} to learn a better generalized feature aggregator, in each GC layer, we concatenate the features before and after the feature propagation and followed by one fully connected layer. 

In our experiments, the output channels are set to $512$ for all layers, and $3$ GC layers are included in each GCN branch. We use the original implementation of HRNet \cite{sun2019hrnet} as the 2D pose estimator and extract PAF from original OpenPose \cite{cao2019openpose}. In addition, we randomly crop the input image by $[0, 0.5w]$ to simulate the cases, where persons are out-of-frame ($w$ is the long side of the cropped image). The GCN is trained for 100 epochs with the Adam \cite{kingma2014adam} optimizer. Learning rate is set to $1e-3$ in beginning and scaled by $0.1$ every $40$ epochs. The training takes about $18$ hours on single Nvidia RTX 2080Ti GPU. 

\subsection{Temporal Convolutional Network}

%Our 2D pose estimator adopts the HRNet structure~\cite{sun2019hrnet} with 4 scales and 4 stages in total. Each stage consists 6 residual blocks and 2 fusion operations. The appearance similarity network (ASN) backbone uses similar structure but with only 3 residual blocks and 1 fusion operation for each stage Furthermore, in ASN, we change the stride of first two convolutional layers from 2 to 1 as down-sampling might cause loss of spatial information. We use maximum of 5 templates for each person. Following training settings used in previous methods~\cite{arnab2019exploiting,kolotouros2019learning,Moon_2019_ICCV_3DMPPE}, our network is trained with multiple 2D datasets including MS-COCO, MPII, and PoseTrack2018, where interpolation is conducted for keypoints to match the keypoint definition of MPII.
%We perform random sampling for training batches and apply hard sample mining in the training process. Adam optimizer is used for training with learning rate from $1e-4$ in beginning to $1e-6$ at the end with exponential decay. Training takes 10000 iterations and about 2 days on 4x2080Ti GPUs.

Our TCNs for estimating the person-centric pose (joint-TCN), depth (root-TCN), and velocity (velocity-TCN) all have 4 residual blocks with dilation rate of 3,9,27,81, respectively. The temporal window length is set to 243 for all our TCNs in all experiments.
We only use Human3.6M dataset for training the TCNs for fair comparison with most previous methods; however, there are a few methods ~\cite{arnab2019exploiting,kolotouros2019learning,Moon_2019_ICCV_3DMPPE} using additional 3D datasets such as MPI-INF-3DHP~\cite{mehta2017monocular} or LSP~\cite{johnson2010clustered}.

\begin{figure}[t]
    \centering
    \includegraphics[width=\linewidth]{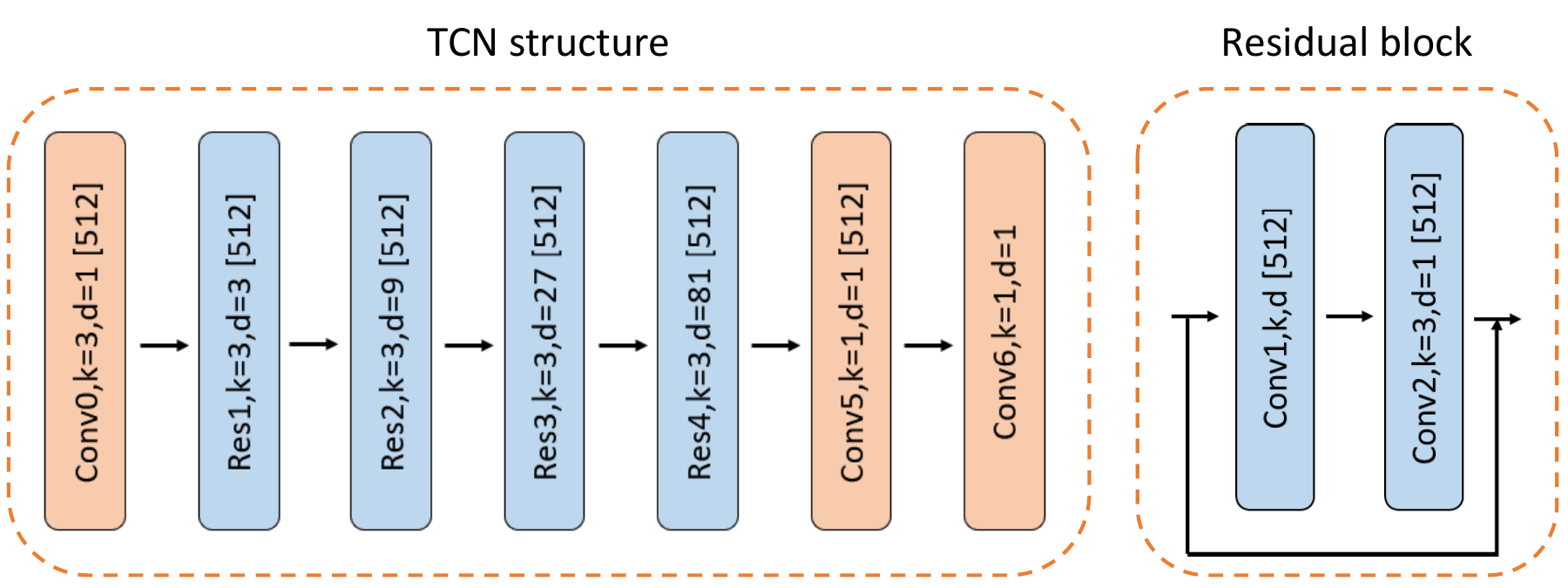}
    \caption{The structure of the TCN. $k$ and $d$ indicates the kernel size and dilation rate, respectively. Number in the bracket $[\cdot]$ represents the number of output channel.}
    \label{fig:tcn}
\end{figure}

We use the same TCN structure (1D convolutional Network) as shown in Fig. \ref{fig:tcn} for the joint-TCN, root-TCN, and velocity-TCN. In the TCN structure, a residual block is repeatedly used and consists of two convolutional layers with skip connections. The first convolutional layer has a kernel size $k$ and dilation rate $d$, the second one has a kernel size $3$ and dilation rate $1$. 

In our experiments, the network is trained with the Adam optimizer \cite{kingma2014adam} for $100$ epochs. The learning rate is set to $1e-3$ in the beginning and scaled by $0.1$ every $40$ epochs. The augmentation method proposed by \cite{cheng2020sptaiotemporal} is used to enhance the robustness to occlusion. The training takes about $40$ hours in single Nvidia RTX 2080Ti GPU.

\begin{figure}[t]
    \centering
    \includegraphics[width=\linewidth]{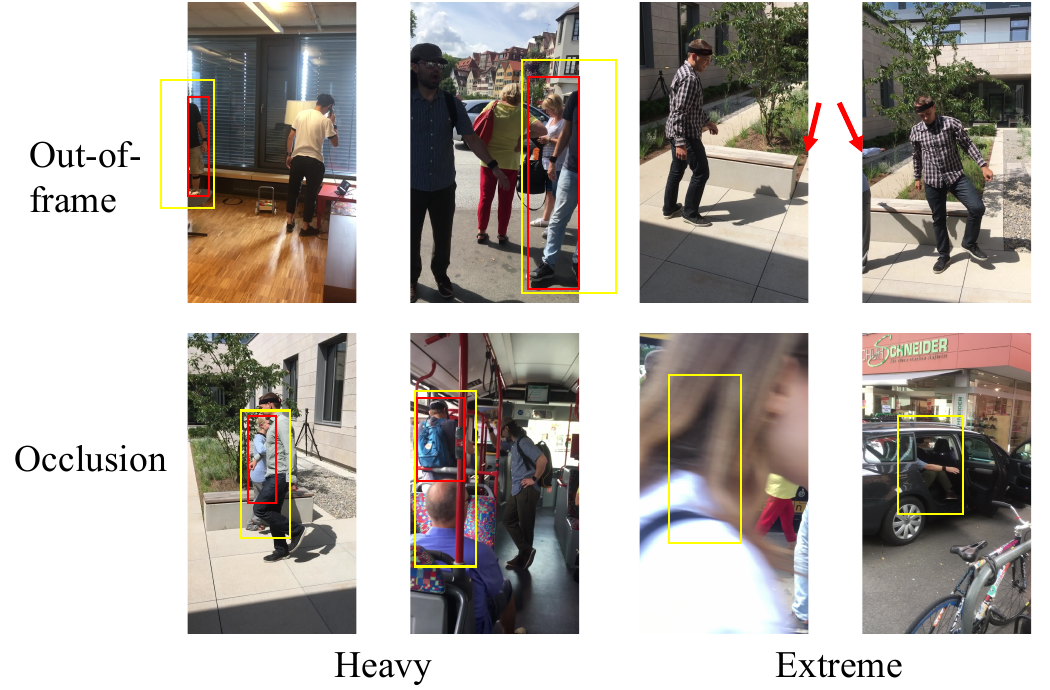}
    \caption{Sample selected frames of heavy and extreme occlusion. Yellow: Ground-truth bounding box. Red: Detection result.}
    \label{fig:3dpwsub}
\end{figure}

\section{Visualization of Selected Subset}

As described in the experiment section of our main paper, we select a subset of 3DPW \cite{3DPW} based on the IoU between the detected bounding box and the ground-truth bounding box from all three sets (train/validation/test) in 3DPW.  Note, 3DPW is  used only for evaluation (not fine-tuning). When computing the IoU, we find that due to the occlusion or out-of-frame problems, the ground truth bounding box of a target person may be incomplete (see examples in Fig \ref{fig:3dpwsub}). When this happens, we choose to re-project the 3D skeleton back to the 2D image plane with the camera parameters provided in the 3DPW dataset, and calculate the ground truth bounding box of the target person based on the re-projected 2D pose. 
%\textcolor{yellow}{The 2D point provided by 3DPW is incomplete when occlusion, so need to calculate the re-projected 3D of ground truth, not for detected results.} 
Fig. \ref{fig:3dpwsub} shows examples of the selected frames with small IoU (half of the body is occluded or out-of-frame) and extreme small IoU (almost entirely out-of-frame or occluded), where the IoU scores are in the range of $[0, 0.6]$. 

\begin{figure}[t]
    \centering
    \includegraphics[width=0.8\linewidth]{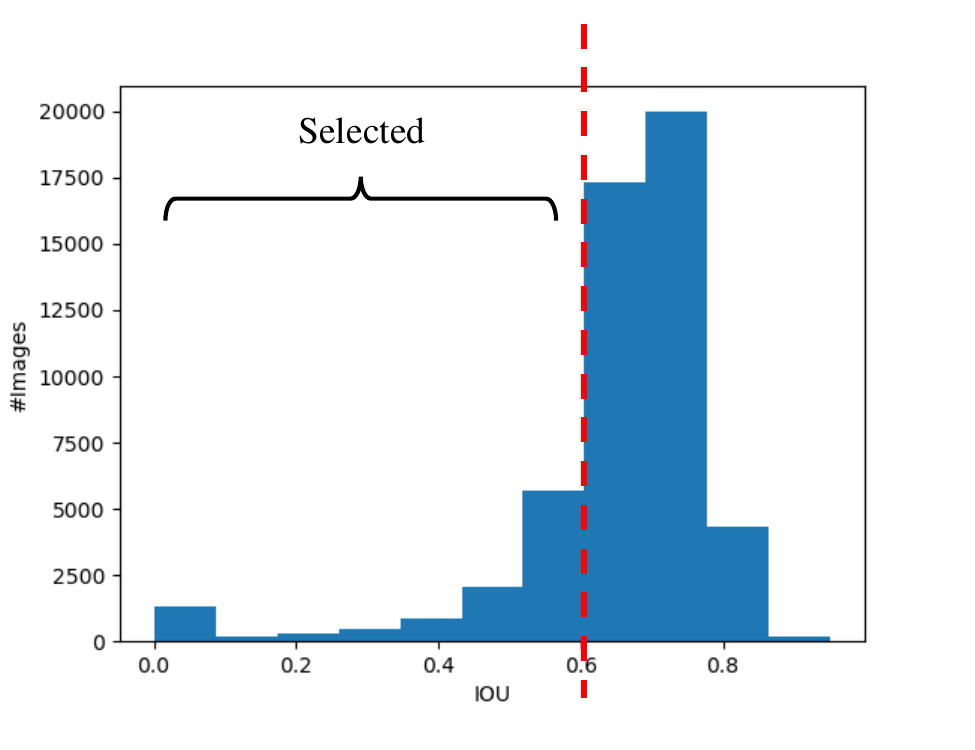}
    \caption{The histogram of IoU between the  detection and ground truth bounding box (reprojected from 3D) on 3DPW. The occlusion subset is chosen from $1^{st}$ to $10000^{th}$ smallest IOU which is $[0, 0.597]$.}
    \label{fig:3dpwhist}
\end{figure}

Fig. \ref{fig:3dpwhist} shows the histogram of IoUs between the detection and ground-truth bounding boxes of all frames in 3DPW. We choose the frames with smallest $1^{st}$ to $10000^{th}$ IoU as the subset, which corresponds to the range $[0, 0.597]$. As we can see that even the IoU histogram peaks around 0.7, there are still many frames where the IoU between the detection bounding box and ground-truth is equal or smaller than $0.6$. As the examples shown in Fig. \ref{fig:3dpwsub}, the detection bounding box with IoU scores in the range of $[0, 0.6]$ is insufficient for reliable 3D human pose estimation. Therefore, it is clear that addressing the missing information problem caused by incorrect detection, out-of-frame or occlusion is critical to reliably estimating 3D multi-person poses.

%\textcolor{blue}{Here, the 0.6 IoU bin contains more than 10000 frames, what is the exact value of the IoU of the 10000th? We can say the 10000th smallest IoU between detected and ground-truth bounding boxes falls into the bin of 0.6 in Fig. \ref{fig:3dpwhist}.} \textcolor{yellow}{The $10000_{th}$ smallest IOU is $0.5973$, which is rounded to 0.6. How to say it more properly? or do I need to re-draw the fig 5?}.

\section{Evaluation Metrics}
Following the literature~\cite{Moon_2019_ICCV_3DMPPE,pavllo20193d}, Mean Per Joint Position Error (MPJPE), Procrustes analysis MPJPE (PA-MPJPE), Percentage of Correct 3D Keypoints (PCK), and area under PCK curve from various thresholds ($AUC_{rel}$) are used to evaluate person-centric 3D human pose estimation. Average precision of 3D human root location ($AP_{25}^{root}$) and PCK$_{abs}$, which is PCK without root alignment to evaluate the absolute camera-centric coordinates, are used to evaluate 3D multi-person camera-centric pose estimation.

\section{Additional Quantitative Results}

\begin{table*}[h!]
    \centering
    \begin{tabular}{cccccccccccc}
        \noalign{\hrule height 1.5pt}
         Method & S1 & S2 & S3 & S4 & S5 & S6 & S7 & S8 & S9 & S10 & - \\ \hline
         \cite{Moon_2019_ICCV_3DMPPE} & 59.5 & 45.3 & 51.4 & 46.2 & \textbf{53.0} & 27.4 & 23.7 & 26.4 & 39.1 & 23.6 & \\ 
         \cite{zhen2020smap} & 42.1 & 41.4 & 46.5 & 16.3 & \textbf{53.0} & 26.4 & \textbf{47.5} & 18.7 & 36.7 & \textbf{73.5} & \\
         Ours & \textbf{64.7} & \textbf{61.1} & \textbf{59.4} & \textbf{63.1} & 52.6 & \textbf{43.1} & 31.9 & \textbf{35.2} & \textbf{53.0} & 28.3 & \\\noalign{\hrule height 1.5pt}
         Method & S11 & S12 & S13 & S14 & S15 & S16 & S17 & S18 & S19 & S20 & Avg \\ \hline
         \cite{Moon_2019_ICCV_3DMPPE}  & 18.3 & 14.9 & 38.2 & 29.5 & 36.8 & 23.6 & 14.4 & 20.0 & 18.8 & 25.4 & 31.8 \\
         \cite{zhen2020smap} & \textbf{46.0} & 22.7 & 24.3 & 38.9 & 47.5 & 34.2 & 35.0 & 20.0 & 38.7 & \textbf{64.8} & 38.7 \\
         Ours & 37.6 & \textbf{26.7} &  \textbf{46.3} & \textbf{44.5} & \textbf{50.2} & \textbf{47.9} & \textbf{39.4} & \textbf{23.5} & \textbf{61.0} & 56.1 & \textbf{46.3} \\\noalign{\hrule height 1.5pt}
    \end{tabular}
    \caption{$PCK_{abs}$ on MuPoTS-3D dataset for matched poses.}
    \label{tab:pckabsm}
\end{table*}

\begin{table*}[h!]
    \centering
    \begin{tabular}{cccccccccccc}
        \noalign{\hrule height 1.5pt}
         Method & S1 & S2 & S3 & S4 & S5 & S6 & S7 & S8 & S9 & S10 & - \\ \hline
         \cite{rogez2017lcr} & 69.1 & 67.3 & 54.6 & 61.7 & 74.5 & 25.2 & 48.4 & 63.3 & 69.0 & 78.1 & \\
         \cite{rogez2019lcr} & 88.0 & 73.3 & 67.9 & 74.6 & 81.8 & 50.1 & 60.6 & 60.8 & 78.2 & 89.5 & \\
         \cite{dabral2018learning} & 85.8 & 73.6 & 61.1 & 55.7 & 77.9 & 53.3 & 75.1 & 65.5 & 54.2 & 81.3 & \\
         \cite{Moon_2019_ICCV_3DMPPE} & \textbf{94.4} & 78.6 & 79.0 & 82.1 & 86.6 & 72.8 & 81.9 & 75.8 & \textbf{90.2} & 90.4 & \\
         \cite{mehta2018single} & 81.0 & 64.3 & 64.6 & 63.7 & 73.8 & 30.3 & 65.1 & 60.7 & 64.1 & 83.9 & \\
         \cite{mehta2017vnect} & 88.4 & 70.4 & 68.3 & 73.6 & 82.4 & 46.4 & 66.1 & 83.4 & 75.1 & 82.4 & \\
         \cite{zhen2020smap} & 89.9 & 88.3 & 78.9 & 78.2 & 87.6 & 54.0 & 88.5 & 71.6 & 70.3 & 89.2 & \\
         Ours & 91.2 & \textbf{90.9} & \textbf{81.7} & \textbf{82.4} & \textbf{88.9} & \textbf{85.0} & \textbf{94.7} & \textbf{91.3} & 81.5 & \textbf{93.2} & \\\noalign{\hrule height 1.5pt}
         Method & S11 & S12 & S13 & S14 & S15 & S16 & S17 & S18 & S19 & S20 & Avg \\ \hline
         \cite{rogez2017lcr} & 53.8 & 52.2 & 60.5 & 60.9 & 59.1 & 70.5 & 76.0 & 70.0 & 77.1 & 81.4 & 62.4 \\
         \cite{rogez2019lcr} & 70.8 & 74.4 & 72.8 & 64.5 & 74.2 & 84.9 & 85.2 & 78.4 & 75.8 & 74.4 & 74.0 \\
         \cite{dabral2018learning} & 82.2 & 71.0 & 70.1 & 67.7 & 69.9 & 90.5 & 85.7 & 86.3 & 85.0 & \textbf{91.4} & 74.2 \\
         \cite{Moon_2019_ICCV_3DMPPE} & 79.4 & 79.9 & 75.3 & 81.0 & 81.0 & 90.7 & 89.6 & 83.1 & 81.7 & 77.3 & 82.5 \\
         \cite{mehta2018single} & 71.5 & 69.6 & 69.0 & 69.6 & 71.1 & 82.9 & 79.6 & 72.2 & 76.2 & 85.9 & 69.8 \\
         \cite{mehta2017vnect} & 76.5 & 73.0 & 72.4 & 73.8 & 74.0 & 89.6 & 84.3 & 73.9 & 85.7 & 90.6 & 75.8 \\
         \cite{zhen2020smap} & 76.3 & 82.0 & 70.8 & 65.2 & 80.4 & 91.6 & 90.4 & 83.4 & 84.3 & 91.2 & 80.5 \\
         Ours & \textbf{86.5} & \textbf{83.1} & \textbf{89.4} & \textbf{91.6} & \textbf{88.1} & \textbf{93.4} & \textbf{90.5} & \textbf{89.5} & \textbf{87.9} & 90.5 & \textbf{88.6} \\\noalign{\hrule height 1.5pt}
    \end{tabular}
    \caption{$PCK$ on MuPoTS-3D dataset for matched poses.}
    \label{tab:pckrelm}
\end{table*}

\begin{table*}[h!]
    \centering
    \begin{tabular}{cccccccccccc}
        \noalign{\hrule height 1.5pt}
         Method & S1 & S2 & S3 & S4 & S5 & S6 & S7 & S8 & S9 & S10 & - \\ \hline
         \cite{Moon_2019_ICCV_3DMPPE} & 59.5 & 44.7 & 51.4 & 46.0 & 52.2 & 27.4 & 23.7 & 26.4 & 39.1 & 23.6 & \\
          \cite{zhen2020smap} & 41.6 & 33.4 & 45.6 & 16.2 & 48.8 & 25.8 & \textbf{46.5} & 13.4 & 36.7 & \textbf{73.5} & \\
         Ours & \textbf{64.7} & \textbf{59.3} & \textbf{59.4} & \textbf{63.1} & \textbf{52.6} & \textbf{42.7} & 31.9 & \textbf{35.2} & \textbf{51.8} & 28.3 & \\\noalign{\hrule height 1.5pt}
         Method & S11 & S12 & S13 & S14 & S15 & S16 & S17 & S18 & S19 & S20 & Avg \\ \hline
         \cite{Moon_2019_ICCV_3DMPPE} & 18.3 & 14.9 & 38.2 & 26.5 & 36.8 & 23.4 & 14.4 & 19.7 & 18.8 & 25.1 & 31.5 \\
         \cite{zhen2020smap} & \textbf{43.6} & 22.7 & 21.9 & 26.7 & 47.1 & 32.5 & 31.4 & 18.0 & 33.8 & 47.8 & 35.4 \\
         Ours & 37.1 & \textbf{23.1} & \textbf{46.3} & \textbf{43.1} & \textbf{50.2} & \textbf{47.9} & \textbf{39.4} & \textbf{21.5} & \textbf{61.0} & \textbf{54.8} & \textbf{45.7} \\\noalign{\hrule height 1.5pt}
    \end{tabular}
    \caption{$PCK_{abs}$ on MuPoTS-3D dataset for all poses.}
    \label{tab:pckabs}
\end{table*}

\begin{table*}[h!]
    \centering
    \begin{tabular}{cccccccccccc}
        \noalign{\hrule height 1.5pt}
         Method & S1 & S2 & S3 & S4 & S5 & S6 & S7 & S8 & S9 & S10 & - \\ \hline
         \cite{rogez2017lcr} & 67.7 & 49.8 & 53.4 & 59.1 & 67.5 & 22.8 & 43.7 & 49.9 & 31.1 & 78.1 &  \\
         \cite{rogez2019lcr} & 87.3 & 61.9 & 67.9 & 74.6 & 78.8 & 48.9 & 58.3 & 59.7 & 78.1 & 89.5 & \\
         \cite{dabral2018learning} & 85.1 & 67.9 & 73.5 & 76.2 & 74.9 & 52.5 & 65.7 & 63.6 & 56.3 & 77.8 & \\
         \cite{Moon_2019_ICCV_3DMPPE} & \textbf{94.4} & 77.5 & 79.0 & 81.9 & 85.3 & 72.8 & 81.9 & 75.7 & \textbf{90.2} & 90.4 & \\
         \cite{mehta2018single} & 81.0 & 59.9 & 64.4 & 62.8 & 68.0 & 30.3 & 65.0 & 59.2 & 64.1 & 83.9 & \\
         \cite{mehta2017vnect} & 88.4 & 65.1 & 68.2 & 72.5 & 76.2 & 46.2 & 65.8 & 64.1 & 75.1 & 82.4 & \\
         \cite{zhen2020smap} & 88.8 & 71.2 & 77.4 & 77.7 & 80.6 & 49.9 & 86.6 & 51.3 & 70.3 & 89.2 & \\
         Ours & 91.2 & \textbf{88.4} & \textbf{81.7} & \textbf{82.4} & \textbf{88.9} & \textbf{84.5} & \textbf{94.7} &  \textbf{91.3} & 78.3 & \textbf{93.2} & \\\noalign{\hrule height 1.5pt}
         Method & S11 & S12 & S13 & S14 & S15 & S16 & S17 & S18 & S19 & S20 & Avg \\ \hline
         \cite{rogez2017lcr} & 50.2 & 51.0 & 51.6 & 49.3 & 56.2 & 66.5 & 65.2 & 62.9 & 66.1 & 59.1 & 53.8 \\
         \cite{rogez2019lcr} & 69.2 & 73.8 & 66.2 & 56.0 & 74.1 & 82.1 & 78.1 & 72.6 & 73.1 & 61.0 & 70.6 \\
         \cite{dabral2018learning} & 76.4 & 70.1 & 65.3 & 51.7 & 69.5 & 87.0 & 82.1 & 80.3 & 78.5 & 70.7 & 71.3 \\
         \cite{Moon_2019_ICCV_3DMPPE} & 79.2 & 79.9 & 75.1 & 72.7 & 81.1 & 89.9 & 89.6 & 81.8 & 81.7 & 76.2 & 81.8 \\
         \cite{mehta2018single} & 67.2 & 68.3 & 60.6 & 56.5 & 59.9 & 79.4 & 79.6 & 66.1 & 66.3 & 63.5 & 65.0 \\
         \cite{mehta2017vnect} & 74.1 & 72.4 & 64.4 & 58.8 & 73.7 & 80.4 & 84.3 & 67.2 & 74.3 & 67.8 & 70.4 \\
         \cite{zhen2020smap} & 72.3 & 81.7 & 63.6 & 44.8 & 79.7 & 86.9 & 81.0 & 75.2 & 73.6 & 67.2 & 73.5 \\
         Ours & \textbf{83.9} & \textbf{80.6} & \textbf{89.4} & \textbf{90.3} & \textbf{88.1} & \textbf{93.4} & \textbf{90.5} & \textbf{87.4} & \textbf{87.9} & \textbf{86.9} & \textbf{87.5} \\\noalign{\hrule height 1.5pt}
    \end{tabular}
    \caption{$PCK$ on MuPoTS-3D dataset for all poses.}
    \label{tab:pckrel}
\end{table*}

%\textcolor{blue}{Hawk: revising this part now.}
The detailed results and comparison with the state-of-the-art (SOTA) methods on MuPoTS-3D dataset \cite{mehta2018single} are shown in Tab. \ref{tab:pckabsm} \ref{tab:pckrelm} \ref{tab:pckabs} \ref{tab:pckrel}. $PCK$ and $PCK_{abs}$ are used as evaluation metrics to measure the person-centric and camera-centric 3D pose estimation accuracy, which are the same as the metrics used in the quantitative results section of our main paper (Table 3). One can observe that our method shows more consistent improvement compared to those of the SOTA methods in all the four evaluations. 

In particular, Tab. \ref{tab:pckabsm} \ref{tab:pckrelm} shows  the $PCK$ and $PCK_{abs}$ evaluations for matched poses and Tab. \ref{tab:pckabs} \ref{tab:pckrel} show the evaluations for all poses, where we follow \cite{mehta2018single,Moon_2019_ICCV_3DMPPE,zhen2020smap}. 

According to Tab. \ref{tab:pckabsm}, we observe that our method performs better than the SOTA methods in 15 out of 20 sequences and in the average score. However, in sequence 10, the size of persons around the image border is abnormal due to camera distortion, which results in the wrong depth estimation of our method. As we use the Human3.6M dataset to train our network, where little distortion exists, the estimation accuracy of our method is negatively affected by this image distortion.

In the Tab. \ref{tab:pckrelm}, we observe a constant improvement of the estimation accuracy in the most sequences (17 out of 20), except for sequence 9 compared to \cite{Moon_2019_ICCV_3DMPPE}. The estimation error is caused by image distortion as well. Different from sequence 10, where the person moves towards the center of the image, resulting in the higher camera-centric estimation error for $PCK_{abs}$, in sequence 9, the person is constantly standing at the distortion area yielding a higher person-centric pose estimation error for $PCK$ of our method.

For more comprehensive evaluation, we also show the results for all poses, where missing poses are counted as wrong in Tab. \ref{tab:pckabs} and \ref{tab:pckrel}. The performance of each method drops as missing detection are counted as wrong for all joints belonging to the target person. Our method maintains high performance while other methods drop significantly due to misdetections. Noticeably, since the $PCK$ and $PCK_{abs}$ do not punish the false positives, \cite{Moon_2019_ICCV_3DMPPE} includes many false-positive bounding boxes, and  thus their method shows constant performance for both matched and all poses.

%\textcolor{red}{Add some description of results; Also, need to explain why we do not include Li et al. and Lin et al. papers into comparisons in Table 1 - 4, because they either did not release their code or their trained model on MuPoTS-3D.}. 

\begin{figure}[h!]
    \centering
    \includegraphics[width=\linewidth]{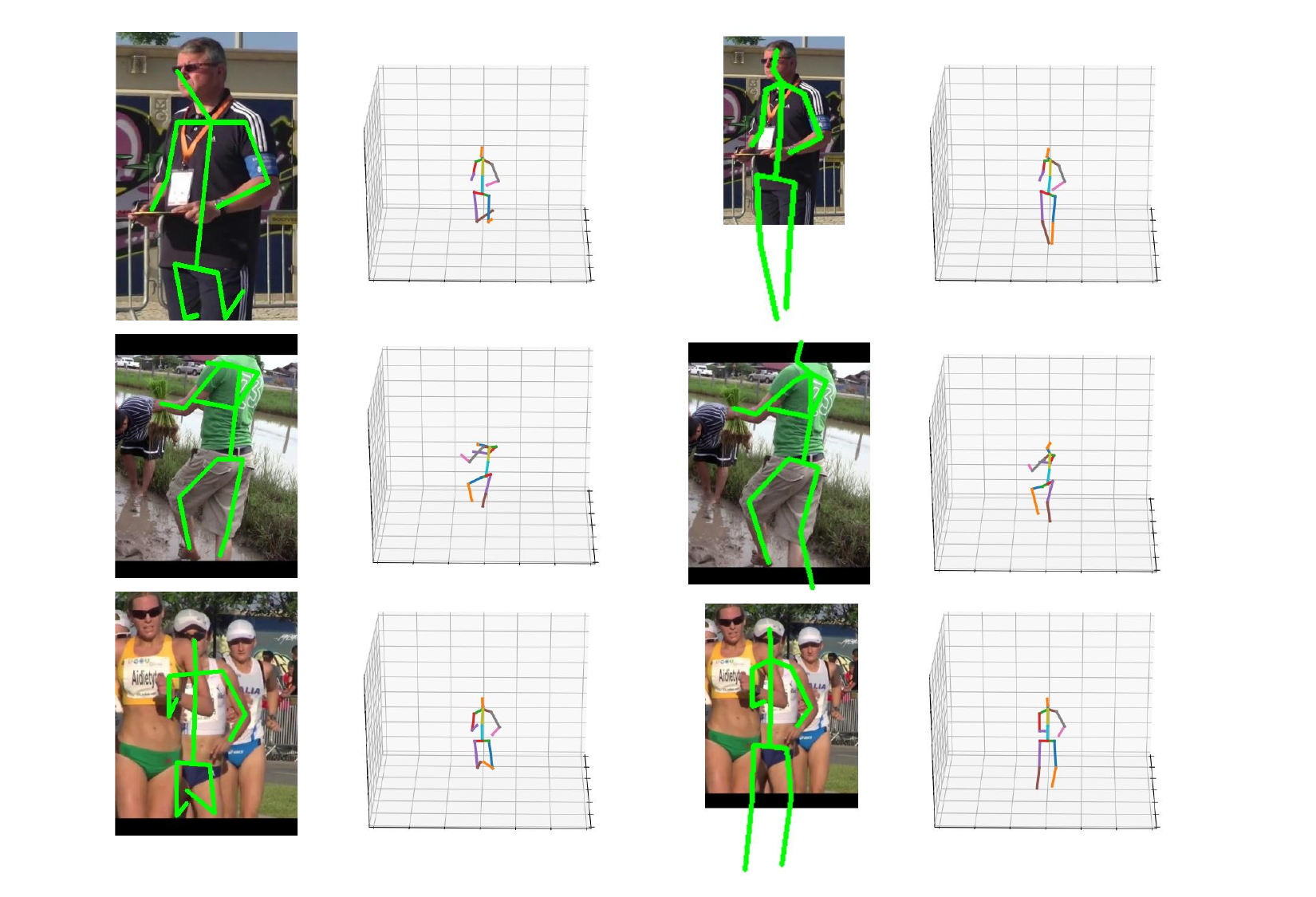}
    \caption{Results of our GCN module in both the 3D human pose estimation and reprojected  2D poses back to the image space. Left two columns: the existing top-down 3D human pose estimation results; right two columns: our results.}
    \label{fig:gcnresult}
\end{figure}

\section{Additional Qualitative Results}

\begin{figure*}[h!]
    \centering
    \includegraphics[width=\textwidth]{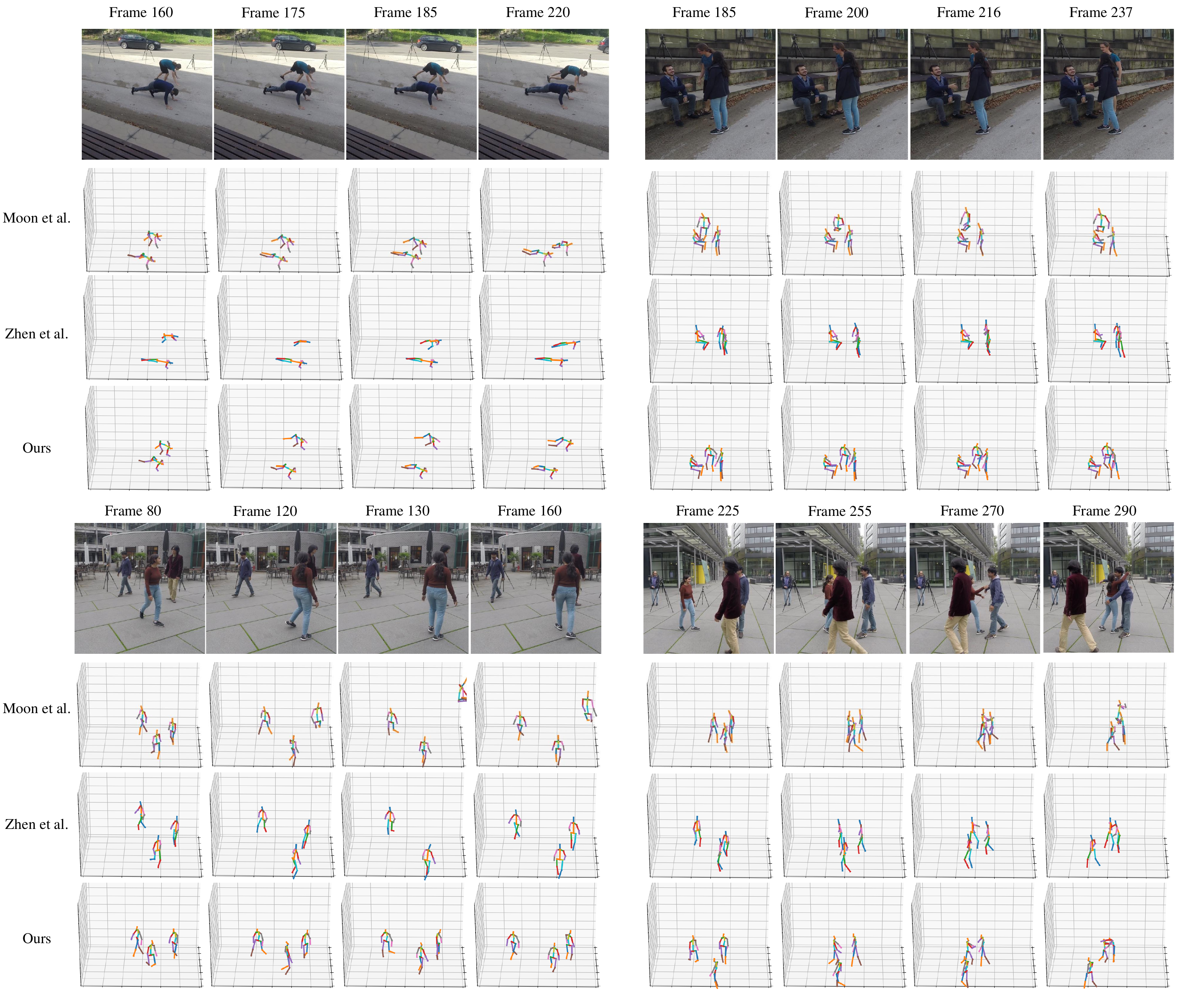}
    \caption{Qualitative comparison with the SOTA 3D multi-person camera-centric pose estimation methods on MuPoTS-3D dataset.}
    \label{fig:compare}
\end{figure*}

To compare with the SOTA methods visually, we run the open-source code of the recent methods \cite{Moon_2019_ICCV_3DMPPE,zhen2020smap}. The 3D multi-person pose estimation results of each method on the frames where occlusion occurs are shown in Fig. \ref{fig:compare}. Since other methods only use frame-wise information, their performance suffers seriously from occlusions, while our temporal-based method produces more robust and accurate pose estimation results.

\begin{figure}[h!]
\centering
\includegraphics[width=0.9\linewidth]{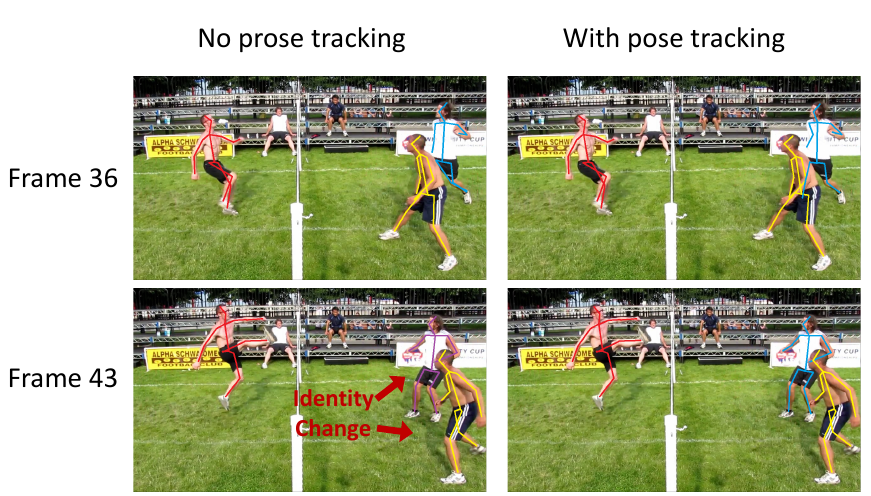}
\caption{Pose estimation results with and without performing pose tracking. Same color indicates the same identity.}
\label{fig:pose_tracking}
\end{figure}

\begin{figure*}[h!]
    \centering
    \includegraphics[width=\textwidth]{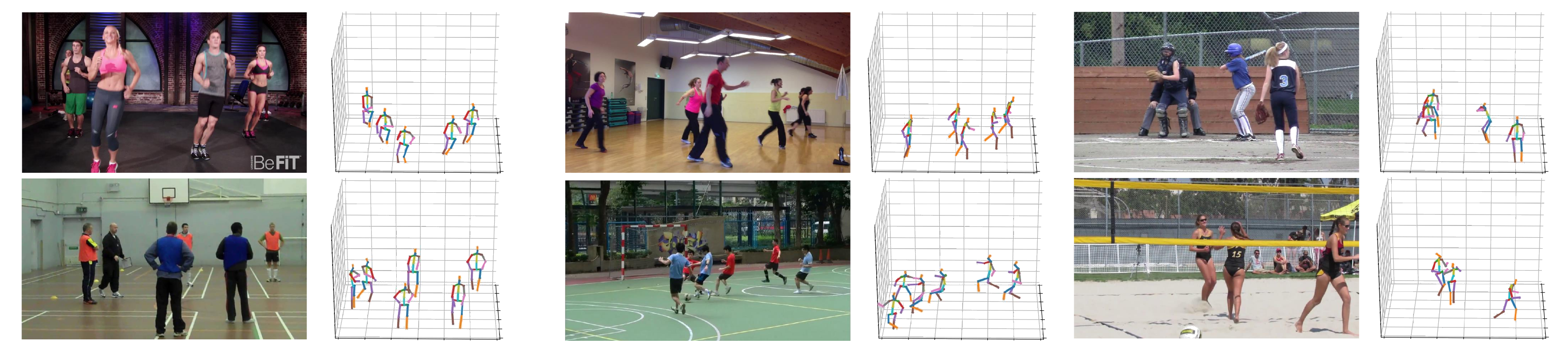}
    \caption{3D camera-centric multi-person pose estimation results of our method on the wild videos (PoseTrack dataset) with occlusions.}
    \label{fig:wild}
\end{figure*}

To illustrate that our method is able to deal with the  out-of-frame problem, we show three cases where the wrongly estimated poses are caused by out-of-frame and occlusion in Fig. \ref{fig:gcnresult}. In a top-down 3D pose estimation framework, once the out-of-frame problem happens, if maximum responses in 2D heatmap are chosen to generate the 3D poses, large error will be produced; since the true joints are out of the image boundary. With our GCNs, the full-body poses are inferred, which is not constrained by the image size. The reprojected the 2D pose from the 3D pose estimation on the original image space is provided to better visualize the effectiveness of our method.

Additionally, we show the estimation results for in-the-wild videos as in Fig. \ref{fig:wild}. Our method can produce reasonable predictions when occlusion occurs.

\section{Pose Tracking illustration}
% Probably still need this part as we barely mention how to do pose tracking but only provide a reference, we can provide some details here to explain how to do it in case the reviewers ask us about the details how to do it. If we do it, we can save some space in our rebuttal.

To perform human pose tracking, we follow~\cite{umer2020self} to evaluate the probability of assigning a joint to an existing or a newly appearing person. In particular, the pose tracking module considers both appearance similarities with accumulated templates, image patches for the target person in the previous frames, and the motion smoothness according to the predicted camera-centric coordinates. 
% As keypoints from other nearby persons are masked out, the risk of estimation error or identity switch is reduced. 
In Figure \ref{fig:pose_tracking}, we can see that without using pose tracking, the identity change problem is shown in the first column, while using pose tracking can fix this problem.

\section{Failure Cases}

\begin{figure}[h!]
    \centering
    \includegraphics[width=\linewidth]{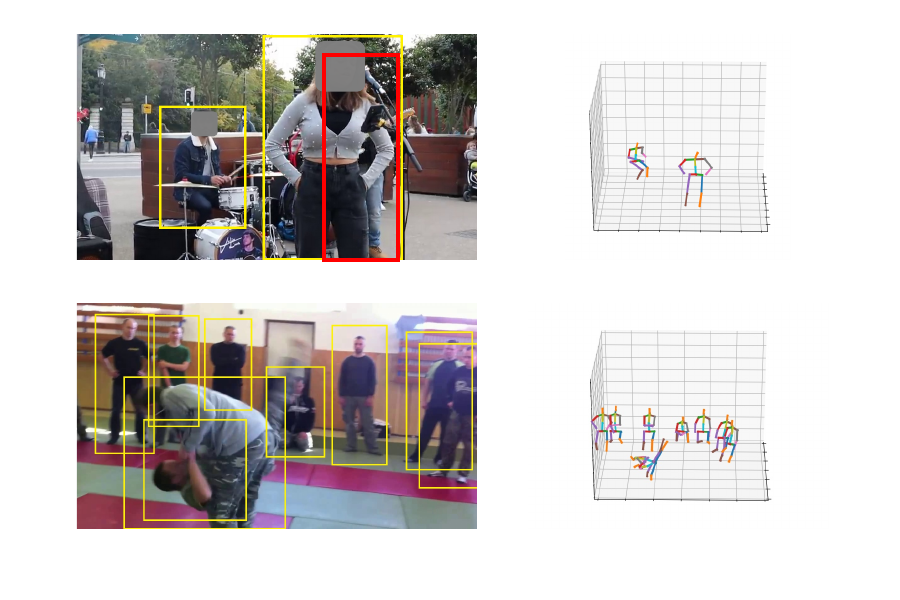}
    \caption{Failure cases. Upper: false human detection, miss detection is highlighted with red bounding box. Lower: unusual pose. Face regions are masked in wild video (upper example), video clip in lower example is from MPII dataset.}
    \label{fig:failure}
\end{figure}

Typical failure cases of our method are shown in Fig. \ref{fig:failure}. As our method belongs to the top-down human pose estimation approach, 
% once the human detector consistently fails in a video clip, the estimated 3D poses of our method are corrupted (the top images). 
severe false human detection (i.e., consistently missing or duplicated) is likely to affect the performance of our method (the upper example in Fig. \ref{fig:failure}).
Moreover, due to the data scarcity of 3D ground-truth (i.e., limited pose variations), our method like many others may not generalize well on wild video. In particular, as our 3D pose estimation is trained only on Human 3.6M dataset \cite{h36m_pami}, which has limited poses variations, once unusual poses occur in the testing video (the lower example in Fig. \ref{fig:failure}), our method may not be able to generalize well. 

%{\small
% \bibliographystyle{aaai21.bst}
%\bibliography{egbib}
%}